\documentclass[11pt, a4paper, logo, copyright]{amd_instella}

\usepackage[authoryear, sort&compress, round]{natbib}
\usepackage{dsfont}
\usepackage{amsmath}
\newcommand{\grayrow}{\rowcolor[gray]{.95}}

\newlength\savewidth

\def\pz{{\phantom{0}}}

\usepackage[utf8]{inputenc} 
\usepackage[T1]{fontenc}    
\usepackage{hyperref}       
\usepackage{url}            
\usepackage{booktabs}       
\usepackage{amsfonts}       
\usepackage{nicefrac}       
\usepackage{microtype}      
\usepackage{xcolor}         

\usepackage{bbm}
\usepackage{xspace}
\usepackage{enumitem}
\usepackage{colortbl}
\usepackage{graphicx, amsmath, amssymb, caption, multirow, overpic, textpos}
\usepackage{floatflt}
\usepackage{subcaption}
\usepackage{algorithmic}
\usepackage{algorithm}

\usepackage{multirow}
\usepackage{wrapfig}

\renewcommand{\today}{2025-6}

\title{Instella-T2I: Pushing the Limits of 1D Discrete Latent Space Image Generation}
\correspondingauthor{Correspondence to: Ze Wang <ze.wang@amd.com>.}

\author{Ze Wang, Hao Chen, Benran Hu, Jiang Liu, Ximeng Sun, Jialian Wu, Yusheng Su, Xiaodong Yu, Emad Barsoum, Zicheng Liu\\
AMD GenAI\\
}

\begin{abstract}
\includegraphics[width=\textwidth]{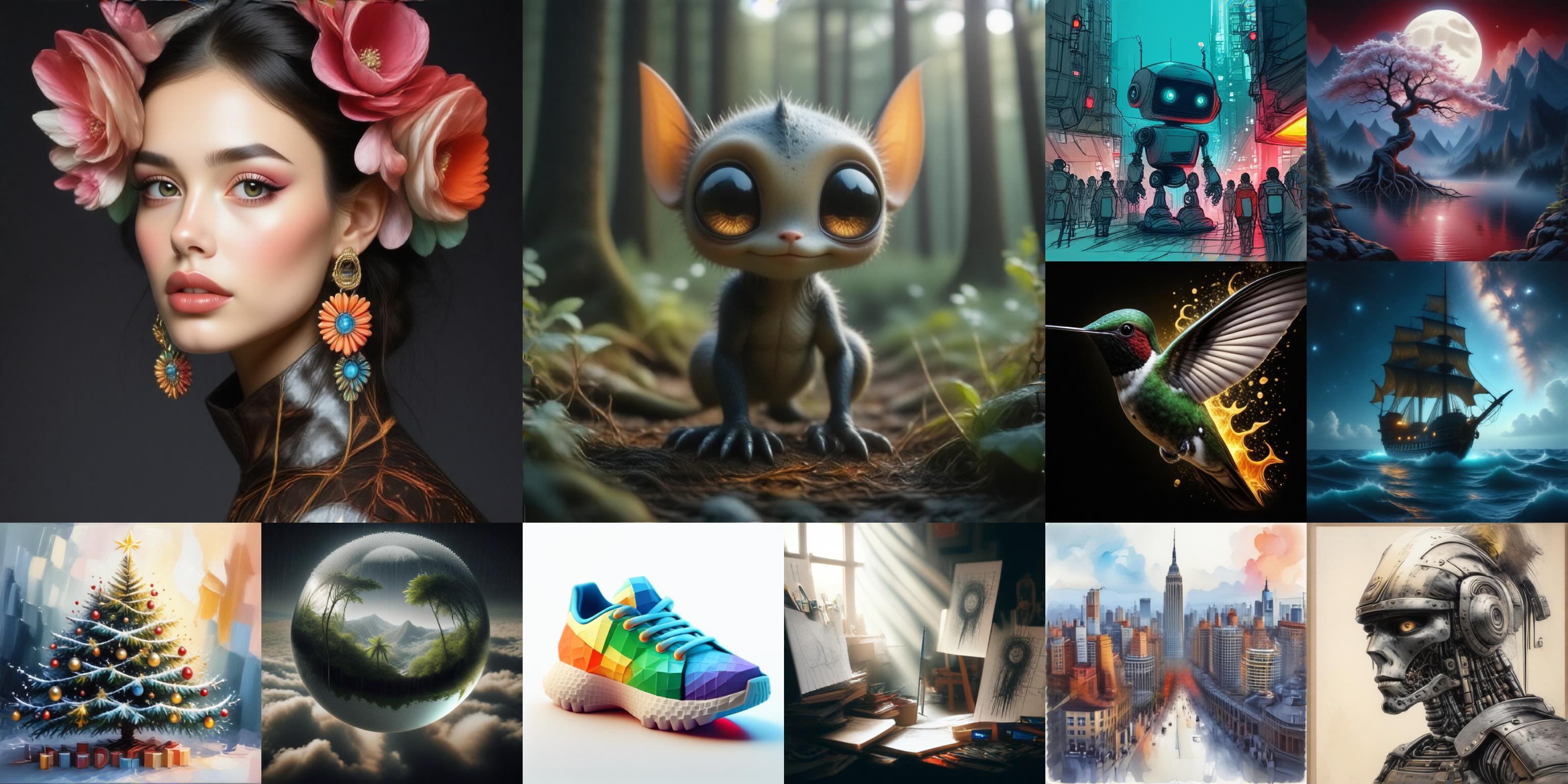}
\captionof{figure}{Text-to-image diffusion and auto-regressive generation samples in 1D binary latent space. Details about the prompts, models, and inference configurations are listed in Appendix Section~\ref{prompts}.}
\label{fig:teaser}
\vspace{5mm}
Image tokenization plays a critical role in reducing the computational demands of modeling high-resolution images, significantly improving the efficiency of image and multimodal understanding and generation. Recent advances in 1D latent spaces have reduced the number of tokens required by eliminating the need for a 2D grid structure. In this paper, we further advance compact discrete image representation by introducing 1D binary image latents.
By representing each image as a sequence of binary vectors—rather than using traditional one-hot codebook tokens, our approach preserves high-resolution details while maintaining the compactness of 1D latents. 
To the best of our knowledge, our text-to-image models are the first to achieve competitive performance in both diffusion and auto-regressive generation using just 128 discrete tokens for images up to $1024\times1024$, demonstrating up to a 32-fold reduction in token numbers compared to standard VQ-VAEs. 
The proposed 1D binary latent space, coupled with simple model architectures, achieves marked improvements in speed training and inference speed. Our text-to-image models allows for a global batch size of 4096 on a single GPU node with 8 AMD MI300X GPUs, and the training can be completed within 200 GPU days. Our models achieve competitive performance compared to modern image generation models without any in-house private training data or post-training refinements, offering a scalable and efficient alternative to conventional tokenization methods.

\end{abstract}

\begin{document}

\maketitle

\section{Introduction}
\label{sec:intro}

Image tokenizers provide compact representations for images, reducing the costs of operating models directly in the high-dimensional raw pixel space \cite{rombach2022high,esser2021taming}. 
With its efficacy in drastically improving the modeling efficiency, image tokenization plays an increasingly crucial role in modern image/multimodal understanding and generation and has been attracting more research attention recently. 
Modern image tokenizers can generally be summarized as continuous and discrete tokenizers, with the Kullback-Leibler variational auto-encoder (KL-VAE) \cite{kingma2013auto} and the vector quantized variational auto-encoder (VQ-VAE) \cite{van2017neural,razavi2019generating,esser2021taming} serving as representatives of each type, respectively.
The KL-VAE and its variants have shown significant promise in image generation.
These models utilize a continuous latent space that effectively encodes rich representations with as few as four channels per image patch \cite{esser2021taming}. 
Meanwhile, the Gaussian distribution of the latent space, imposed by a KL divergence, facilitates a simpler modeling of the latent representations. 
KL-VAE dominates image tokenization for modern diffusion-based image generation methods \cite{rombach2022high,stablediffusion,chen2023pixart,podell2023sdxl,peebles2023scalablediffusionmodelstransformers,ma2024sit,li2024autoregressive}, where progressive denoising of diffusion models can precisely regress latent values and generate visually appealing images (and videos) under various conditions. 
However, the continuous latent space challenges the applications of KL-VAE in auto-regressive image generation, where the `one-shot' token prediction may not be able to regress the latent values at a desired precision, leading to poor image generation quality requiring further refinements such as diffusion loss \cite{li2024autoregressive,fan2024fluid}.

VQ-VAE and its variants \cite{yu2021vector,mentzer2023finite,wang2023binary,tian2024visualautoregressivemodelingscalable,kumar2024high,yu2024language,zhao2024image} provide a discrete latent space with a categorical distribution, which is well-suited for modeling with auto-regressive models \cite{li2023magemaskedgenerativeencoder,chang2022maskgitmaskedgenerativeimage,sun2024autoregressive,tian2024visualautoregressivemodelingscalable}. 
This discrete nature makes VQ-VAE particularly appealing for multi-modal image understanding and generation, as it can be seamlessly integrated into large language models (LLMs) \cite{radford2019language,brown2020language,gpt4,llama}. 
However, VQ-VAE training is prone to codebook collapse \cite{ramesh2021zero,esser2021taming,huh2023straightening}, which requires additional loss terms and careful hyper-parameter tuning to mitigate \cite{zhu2024addressing,fifty2024restructuring}. More critically, representing each image patch as a single vector selected from a learned latent codebook limits the representation capacity of the latent space \cite{takida2022sq,chen2024softvq}. This constraint necessitates a large number of latent tokens and an expansive codebook to handle high-resolution images \cite{razavi2019generating}. 
For instance, the VQ-VAE model used in EMU3 \cite{wang2024emu3} employs 4096 tokens to represent a single 1024$\times$1024 image, resulting in an extremely slow generation speed of $\sim$5 minutes per image, preventing the scalability to real-world applications.

Traditional latent image representations using 2D latent grids allocate uniform capacity across all image patches \cite{van2017neural}. 
Recent research \cite{yu2024an,chen2024softvq} has shown that this gridded approach can introduce significant redundancy, which, in turn, limits the reconstruction quality of regions with complex features. 
By learning 1D latent space representations \cite{yu2024imageworth32tokens,li2024imagefolder,zha2024language,chen2025masked}, the spatial correspondence between the latent space and the image patches is removed, enabling more compact latent representations that can achieve comparable or even superior reconstruction quality. However, training 1D tokenizers, particularly with discrete latent spaces, can present practical challenges and often requires complex model architectures and training strategies \cite{kim2025democratizing}, such as two-stage training \cite{yu2024imageworth32tokens}, due to its training cost and learning difficulty.
Additionally, the use of discrete 1D tokenizers for high-resolution image generation remains significantly underexplored.

In this paper, we strive to further push the limit of 1D image tokenizers in discrete latent space for fast and high-fidelity image generation with various conditions. We specifically focus on generating higher-resolution images than in previous studies on 1D discrete image tokenizers while utilizing the same or fewer latent tokens.
To increase the representational capacity of each discrete image token, we adopt a binary latent space \cite{bnl} approach, representing each token as a binary vector rather than a one-hot selection from the codebook. This method allows each latent token to be expressed as a binary combination of elements of a learned codebook, significantly expanding its representational capacity and paving the way for a more compact latent space \cite{yu2024language}. We train the binary 1D image tokenizer to decode images to multiple resolutions given the same image latent, allowing for flexible image generation and high-reolution images to be generated at the same high speed as the lower-resolution ones.
For text-to-image generation, we use a pretrained LLM as the text feature encoder and introduce a streamlined but powerful image generation model that jointly processes text tokens extracted from intermediate layers of the LLM across Transformer blocks with image tokens using self-attention. 

We conduct experiments on text-to-image generation with both diffusion and auto-regression models. Our approach allows a 1024$\times$1024 image to be represented with only 128 discrete tokens. Furthermore, we demonstrate that our binary diffusion model in a 1D latent space efficiently supports text-to-image generation, requiring as few as 20 denoising steps without the need for post-training timestep distillation. Combining the reduced number of tokens and the denoising steps, our 1B text-to-image model can generate an image in under 0.5 seconds on a single GPU. Without extensive data curation and post-training, our text-to-image models achieve competitive results with a CLIP score \cite{hessel2021clipscore} of 0.332, an ImageReward score \cite{xu2023imagereward} of 0.90, and an avarage GenEval\cite{ghosh2024geneval} scaore of 0.64 after pretraining.
Moreover, the compact latent space of our model enables a 1B text-to-image diffusion/AR model to be trained on a single 8-GPU node with a global batch size of 4096 for both the resolution of 512 and 1024, achieving a high throughput of approximately 120 images per second per GPU. This efficiency makes text-to-image pretraining feasible within just 20 days on a single computation node (approximately 200 GPU-days). 

\newcommand{\enc}{\mathcal{E}}
\newcommand{\dec}{\mathcal{D}}
\newcommand{\qtz}{\mathcal{Q}}
\newcommand{\img}{\mathbf{x}}
\newcommand{\latent}{\mathbf{z}}
\newcommand{\R}{\mathbb{R}}

\section{Method}
\label{sec:method}

In this section, we present the details of the proposed 1D binary image tokenizer and the accompanying binary diffusion and auto-regressive models.

\begin{figure*}[t]
    \centering
	\includegraphics[width=\linewidth]{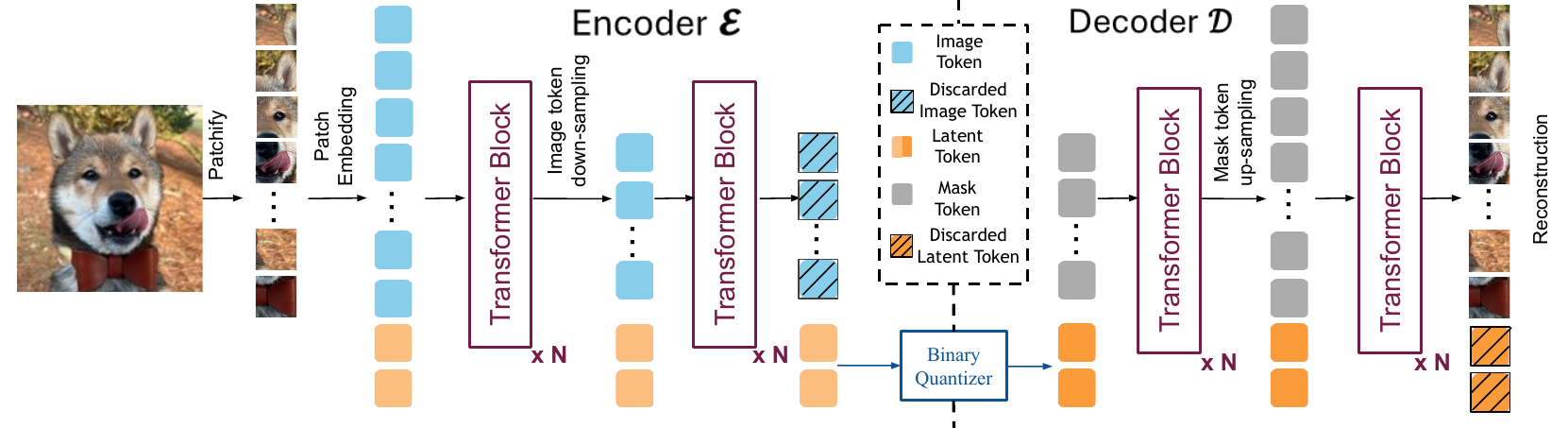}
    \caption{Model architecture of 1D binary image tokenizers (BIT). BIT consists of an encoder $\mathcal{E}$, a quantizer $\mathcal{Q}$, and a decoder $\mathcal{D}$. $\mathcal{E}$ takes both image tokens and latent tokens as input, $\mathcal{Q}$ quantizes the updated latent tokens from the encoder as binary codes, which is fed into $\mathcal{D}$ to reconstruct the original images from mask tokens. For more efficient modeling, we use down/up-sampling layers at the encoder and decoder to compress and expand the image tokens and mask tokens, respectively.}
\label{fig:bae}
\end{figure*}

\subsection{1D Binary Image Tokenizer}

The proposed 1D binary image tokenizer (1D-BIT) adopts the Vision Transformer \cite{dosovitskiy2020image,yu2021vector} architecture and consists of three main components: the encoder $\enc$, the decoder $\dec$, and the quantizer $\qtz$. The architecture of 1D-BIT is illustrated in Figure~\ref{fig:bae}.

\paragraph{Encoder and decoder.} The Transformer-based encoder of 1D-BIT directly processes images as input. Given an image $\img \in \R ^{h \times w \times 3}$, a patch embedding layer first converts the image into a sequence of image tokens $\mathbf{x}_\text{p} \in \R^{l\times c}$ with a patch size of $p$, where $l = hw / p^2$ and $c$ is the hidden dimension size. 
The encoder then applies Transformer blocks jointly on the concatenated sequence of image tokens $\mathbf{x}_{\text{p}}$ and latent tokens $\mathbf{z} \in \R^{k \times c}$. Through joint self-attention layers, latent tokens progressively extract and compress features from both the image tokens and latent tokens themselves. The resulting latent tokens $\mathbf{z}$, which contain the compressed image information in significantly fewer tokens than the original image tokens, are then passed into a binary quantizer (described later) for latent quantization. After encoding, the image tokens are discarded, leaving only the quantized 1D binary latent tokens, represented as a sequence of binary vectors, to serve as the latent representation of the image.

To reconstruct the original image, we learn in the decoder a mask token, which is replicated into a sequence of tokens $\mathbf{m}_{\text{p}} \in \R^{l \times c}$, each corresponding to an image patch, to match the desired latent size and concatenated with the binary latent tokens to serve as the input of the decoder Transformer layers. 
The decoder architecture mirrors that of the encoder, applying Transformer blocks to allow mask tokens to progressively retrieve image features from the latent tokens: 
\begin{equation}
\mathbf{z} = \enc([\mathbf{x}_\text{p}; \mathbf{x}_{z}]), \quad
\tilde{\latent} = \qtz(\mathbf{z}), \quad
\mathbf{m} = \dec([\mathbf{m}_{\text{p}}; \tilde{\latent}]), 
\end{equation}
After processing through the Transformer blocks, the updated latent tokens are discarded and the updated mask tokens $\mathbf{m}$ are used to reconstruct the image.
Specifically, a pixel prediction head $\mathcal{P}$ is applied to the updated mask tokens to regress the values of the image pixels: $\tilde{\mathbf{x}} = \mathcal{P}(\mathbf{m})$. During training, the reconstructed image $\tilde{\mathbf{x}}$ is compared with the original image $\mathbf{x}$ to calculat the loss.
While previous studies \cite{yu2021vector,li2024imagefolder,chen2024softvq} have demonstrated that a simple linear layer can achieve satisfactory reconstruction quality, we find that replacing the linear head with a lightweight convolutional decoder—consisting of transpose convolution, normalization, and ReLU layers—significantly enhances reconstruction quality, particularly for detailed content such as text and human faces. We leave the exploration of a hybrid architecture incorporating these improvements into 1D-BIT for future research.
For future extensions to the support of arbitrary resolution and aspect ratio, we apply rotary position embedding (RoPE) to encode the spatial position of image tokens. 

\paragraph{Token up/down-sampling.} Following common practice, we adopt a consistent patch size of $16\times16$ in all experiments, representing each $512\times512$ image and  $1024\times1024$ as $32\times32$ and $64\times64$ image patches, respectively. However, the resulting long sequence of image patches can reduce efficiency during both training and inference, especially when scaling to higher image resolutions. To address this issue, we insert image downsampling and upsampling layers into the encoder and decoder, respectively, thereby reducing the number of intermediate tokens processed by the Transformer blocks. Specifically, during encoder downsampling, we rearrange the sequence of image tokens into a 2D grid and apply a linear layer to merge each $2\times2$ token group into a single image token:
\begin{equation}
\begin{aligned}
    \mathbf{x}'_{i,j} = \left[
\mathbf{x}_{2i,2j};\; \mathbf{x}_{2i+1,2j};\; \mathbf{x}_{2i,2j+1};\; \mathbf{x}_{2i+1,2j+1}\right],  \quad
\hat{\mathbf{x}}_{i, j} = W_d \mathbf{x}'_{i, j}
\end{aligned}
\end{equation}
where $W_d \in \mathbb{R}^{c \times 4c}$ is the weight of the linear layer and $\hat{\mathbf{x}} \in \R^{\frac{l}{4} \times c}$.
Similarly, for up-sampling in the decoder, we apply a linear layer to expand the hidden dimension of each mask token by $4\times$ and reconstruct each $2\times 2$ token group by splitting the expanded tokens:
\begin{equation}
\begin{aligned}
    \mathbf{m}'_{i,j} = W_u\, \mathbf{m}_{i,j}, \quad \hat{\mathbf{m}}_{i,j} = \operatorname{reshape}\Big(\mathbf{m}'_{i,j},\,(2,2,c)\Big),
\end{aligned}
\end{equation}
where $W_u \in \mathbb{R}^{4c \times c}$ is the weight of the linear layer and $\hat{\mathbf{m}} \in \R^{4l \times c}$.
Note that, unlike convolution-based image tokenizers, we maintain a consistent hidden dimension across all layers during downsampling. Additionally, the number of latent tokens remains unchanged throughout the upsampling and downsampling operations.

\paragraph{Binary quantizer.} In order to obtain the discrete binary latent representation, we apply binary quantization to the latent representations between $\enc$ and $\dec$. 
Following \cite{bnl}, in the binary quantizer, we first apply a linear layer $\mathcal{R}$ to reduce the latent dimension from $c$ to $\hat{c}$. We use Sigmoid activation (denoted as $\sigma(\cdot)$) to normalize the latent to $[0, 1]$ and Bernoulli sampling (denoted as $\mathcal{B}(\cdot)$) to obtain the final discrete representation $\hat{\latent}$ of the image. Specifically, $\tilde{\latent}$ is obtained as:
\begin{equation}
    \tilde{\latent} = \mathcal{B}(\sigma(\mathcal{R}(\latent) / \tau)),
\end{equation}
where $\tau$ denotes the temperature for sampling. The value of $\tau$ is adjusted across different stages of training and is set to $0$ (with no stochasticity) during inference. The stochastic sampling of the latent representation in training effectively prevents code collapse, ensuring that each element in the binary latent space carries valid information and does not collapse to a consistent value across all images. The resulted $\tilde{\latent}$ which contains $k\times\hat{c}$ binary values serves as the latent representation of the image, with $k < l$ and $\hat{c} < c$. 
Note that although our 1D quantization shares a similar form with lookup-free quantization \cite{yu2024language} (LFQ)  at a special variant of binary values ($\{-1, 1\}$ as in LFQ and $\{0, 1\}$ as in binary latent), these two approaches originate from fundamentally different motivations. LFQ models discrete latent as an integer set, whereas our binary latent space models image representation and generation explicitly within a Bernoulli-distributed latent space. Furthermore, the Bernoulli latent representation naturally supports both diffusion and auto-regressive generation methods (as detailed later), while LFQ is exclusively designed for auto-regressive generation.

\paragraph{Multi-stage training.} Training the Transformer-based tokenizer with a discrete quantizer can be empirically challenging, particularly when coupled with adversarial loss \cite{goodfellow2020generative,isola2018imagetoimagetranslationconditionaladversarial}, which, despite its instability, is essential for high-fidelity reconstructions. To balance reconstruction quality and training stability, we adopt a multi-stage training strategy.

In the first stage, the model is trained with images at a resolution of 512$\times$512, using a combination of smooth L1 loss and LPIPS perceptual loss \cite{larsen2016autoencoding,johnson2016perceptual,dosovitskiy2016generating,zhang2018unreasonableeffectivenessdeepfeatures} to update all parameters end-to-end. Specifically, when applying perceptual loss, we resize both input and reconstructed images to $448\times448$ before feeding them into the VGG-based LPIPS module to match global appearance. We also randomly crop $224\times224$ patches from the input and reconstructed images to refine local image details. We consistently observe that replacing smooth L1 loss with other alternatives, such as MSE loss, can lead to loss spikes and even divergence.

In the second stage, we train the model to decode images at multiple resolutions. The output image resolution is controlled by the number of replicated mask tokens. We feed the encoder with images of a consistent size of 512$\times$512, and train the deocder to decode images at various resolutions, including 512$\times$512, 768$\times$768, and 1024$\times$1024. The ROPE layers in the deocder adapt to different resolutions by interpolating the position index of the mask tokens when calculating the rotation angle.

In the third stage, we introduce adversarial training while freezing the parameters of the encoder $\enc$, updating only the decoder and quantizer. The discriminator is built upon a frozen DINO-S model. We observe that the third-stage training is particularly sensitive to the weighting of adversarial loss, prompting us to employ adaptive weighting to balance adversarial and supervised losses. We set $\tau=0.5$ and $\tau=0.1$ for the first and second stages, respectively, where $\tau=0.5$ in the first stage prevents code collapse, and $\tau=0.1$ in the second stage continuously introduces controlled imperfections to the binary codes, thus improving the robustness of image reconstruction.

\subsection{Diffusion in 1D Binary Latent Space}

The diffusion process in our model builds upon the Bernoulli diffusion process introduced in \cite{bnl}. Unlike \cite{bnl}, which employs discrete timesteps, we adopt a continuous time formulation with timesteps in the range of $[0, 1]$. 
Following \cite{esser2024scaling}, we sample training timesteps from a logit-normal distribution to put more weight on training of intermediate denoising steps.
For conciseness, in the following sections, we slightly abuse notation by using $\latent$ rather than $\tilde{\latent}$ to denote the quantized latent representation.
During training, given the binary latent representation $\latent$ of an image, we randomly sample a timestep $t$ from a uniform distribution: $t \sim \mathcal{U}(0, 1)$. The corresponding noisy latent representation is then sampled as: $\latent^t \sim \mathcal{B}(0.5t + (1-t)\latent)$. This distribution can be interpreted as a linear interpolation between a deterministic Bernoulli distribution $\mathcal{B}(\latent)$, where the parameters contain only $0$ and $1$, and a fully stochastic Bernoulli distribution $\mathcal{B}(0.5 \cdot \mathbf{1}_{k \times \hat{c}})$.

Our model $f_\theta$ takes as input the noisy sample $\latent^t$, the timestep $t$, and optionally a conditioning variable $\mathbf{c}$, such as a class label or text prompt. Unless stated otherwise, we adopt an `epsilon’-prediction style prediction target and train the model to predict the flipping probability between the original $\latent$ and the predicted $\hat{\latent}$. This is formualted as: $p(\hat{\latent}) = \sigma(f_\theta(\latent^t, t, \mathbf{c})) \oplus \latent^t$, where $\oplus$ denotes the element-wise logic XOR operation. The model is trained using the binary cross-entropy (BCE) loss: 
\begin{equation}
\mathcal{L}_{\text{diff}} = \mathbb{E}_{t, \latent}\text{BCE}(\sigma(f_\theta(\latent^t, t, \mathbf{c})) \oplus \latent^t, \latent).
\end{equation}
We find this loss function leads to the most stable training dynamics and produces robust results. 

For conditional generation, following \cite{bnl}, we apply both sampling temperature and classifier-free guidance (CFG) \cite{ho2022classifier} in inference as 
\begin{equation}
p(\hat{\latent}) = \sigma((1 + \alpha) f_\theta(\latent^t, t, \mathbf{c}) / \tau - \alpha f_\theta(\latent^t, t, \varnothing) / \tau), 
\end{equation}
where $\alpha$ and $\tau$ are the guidance scale and the temperature, respectively, and $\varnothing$ denotes empty conditions. To support CFG, we randomly drop the input condition $\mathbf{c}$ with a probability of 10\% during training. Apart from the original sampling algorithm introduced in \cite{bnl}, we introduce a simplified sampling algorithm that we observe to perform better with low sampling steps, such as 20 steps. Two sampling algorithms are detailed in Appendix Section~\ref{algo}.

Timestep information is incorporated into the features using adaptive layer normalization layers. Specifically, a linear layer encodes the continuous timestep scalars into a feature vector matching the hidden feature dimensions. Within each transformer block, an adaptive layer normalization zero (adaLN-Zero) layer is learned to modulate the image tokens based on the timestep feature vector.

\subsection{Auto-regressive in 1D Binary latent Space}
While recent advancements, such as vision auto-regressive models \cite{tian2024visualautoregressivemodelingscalable}, have drastically improved the efficiency and performance of auto-regressive image generation, we focus primarily on conventional auto-regressive generation in token-by-token style due to its natural compatibility with modern LLMs, which helps pave the way for further effort on high-efficiency multi-modal unified understanding and generation models with 1D image latent space. 
We leave text-to-image generation with mask decoding style generative models similar to models such as MaskGIT \cite{chang2022maskgitmaskedgenerativeimage} and MUSE \cite{chang2023muse} in furture work. The architecture of AR models closely resembles that of diffusion models, except that there is no timestep embedding layer and the adaLN-Zero layers are replaced with standard layer normalization layers. During training, the prediction of each token is trained by giving text features (denoted as $\mathbf{c}$) and all preceding image tokens as conditions and a loss function: 
\begin{equation}
    \mathcal{L}_\text{AR} = \frac{1}{I} \sum_{i=1}^{I}\text{BCE}(\sigma(f_\theta(\latent[0:i-1], \mathbf{c})), \latent[i]). 
\end{equation}
$\latent[0]$ is a learnable splitting token placed between the text features and the image tokens, and is used for the prediction of $\hat{\latent}[1]$. The prediction of all tokens is learned jointly, leveraging causal attention. During inference, each token is predicted conditioned on the predictions of previous tokens $\hat{\latent}[i] = \mathcal{B}(\sigma(f_\theta(\hat{\latent}[0:i-1], \mathbf{c}) / \tau))$ with $\hat{\latent}[1] = \latent[0]$ and $\tau$ denoting the sampling temperature. Similar to diffusion inference, we can improve the prompt following by using classifier-free guidance as
\begin{equation}
\hat{\latent}[i] = \mathcal{B}(\sigma((1+\alpha)f_\theta(\hat{\latent}[0:i-1], \mathbf{c}) / \tau - \alpha f_\theta(\hat{\latent}[0:i-1], \varnothing) / \tau))
\end{equation}

\section{Related Work}

\paragraph{Discrete latent space.}
Discrete latent space models have been extensively studied in generative modeling. 
VQ-VAEs and their variants have facilitated various auto-regressive generation frameworks. PixelCNN \cite{van2016conditional} was initially applied within the discrete latent space of VQ-VAE \cite{van2017neural,razavi2019generating} for generation.
Later, a token-by-token auto-regressive transformer \cite{esser2021taming} was introduced alongside VQ-VAE, demonstrating faster generation speeds and improved quality over PixelCNN.
Under this auto-regressive framework, different variants of VQ-VAE have been explored, including residual quantization \cite{lee2022autoregressive,kumar2023high} and multi-scale quantization \cite{tian2024visualautoregressivemodelingscalable,li2024imagefolder}. However, this paradigm suffers from two major limitations: codebook collapse in VQ-VAE \cite{huh2023straightening} and error accumulation during auto-regressive inference \cite{li2024imagefolder}.
To address these challenges, various quantization techniques have been proposed, including Finite Scalar Quantization (FSQ) \cite{mentzer2023finite}, Look-up Free Quantization (LFQ) \cite{yu2024language}, Product Quantization (PQ) \cite{li2024imagefolder,qu2024tokenflow}, and Binary Sphere Quantization (BSQ) \cite{zhao2024image}. 
Additionally, several training strategies \cite{zhu2024addressing,fifty2024restructuring} have been developed to enhance the reconstruction and generation performance of VQ-VAE and mitigate its training issues.
On the generative modeling front, alternative auto-regressive paradigms, such as set-by-set \cite{chang2022maskgitmaskedgenerativeimage} and scale-by-scale \cite{tian2024visualautoregressivemodelingscalable} approaches, have been proposed, which enabled faster and higher-quality generation. 
These advances have led to more promising techniques for text-to-image generation models \cite{wu2024liquid,han2024infinity,zhuang2025vargpt}.

\paragraph{1D latent space.}
The concept of 1D latent space was first introduced in SEED \cite{ge2023plantingseedvisionlarge} and TiTok \cite{yu2024imageworth32tokens}, where a set of learnable latent tokens replaces traditional image tokens for quantization and downstream generative modeling. Unlike conventional latent spaces that rely on fixed-grid quantization, this design allows for a flexible number of latent tokens. 
By reducing the number of tokens required for generation, 1D latent space facilitates more efficient autoregressive and diffusion-based models.
However, training 1D latent spaces remains challenging due to the need for effective token representations and stable training dynamics. 
Both SEED and TiTok adopt a two-stage decoder paradigm to mitigate these difficulties. Subsequent works such as ImageFolder \cite{li2024imagefolder} and TokenFlow \cite{qu2024tokenflow} extend the 1D latent space concept through multi-branch and multi-scale quantization schemes, enhancing both efficiency and generation quality. 
SoftVQ-VAE \cite{chen2024softvq} further explores 1D latent tokens in a continuous space with a high compression ratio, demonstrating promising results in diffusion-based models by enabling smoother token transitions.
Recent advances have continued to push the boundaries of 1D latent spaces. 
MAETok \cite{chen2025masked} investigates their potential in pure auto-encoder architectures and masked modeling, eliminating the need for variational posteriors while preserving generation quality. 
Furthermore, text token and text feature alignment within the latent space \cite{zha2024language,kim2025democratizing} have emerged as promising research directions, with the aim of improving cross-modal learning and enabling more effective text-to-image generation.

\section{Text-to-Image Generation}
\label{sec:t2i}

In this section, we further scale up the experiment and present results on text-to-image generation using the proposed 1D binary latent space.


\begin{figure}[t]
    \centering
	\includegraphics[width=1\linewidth]{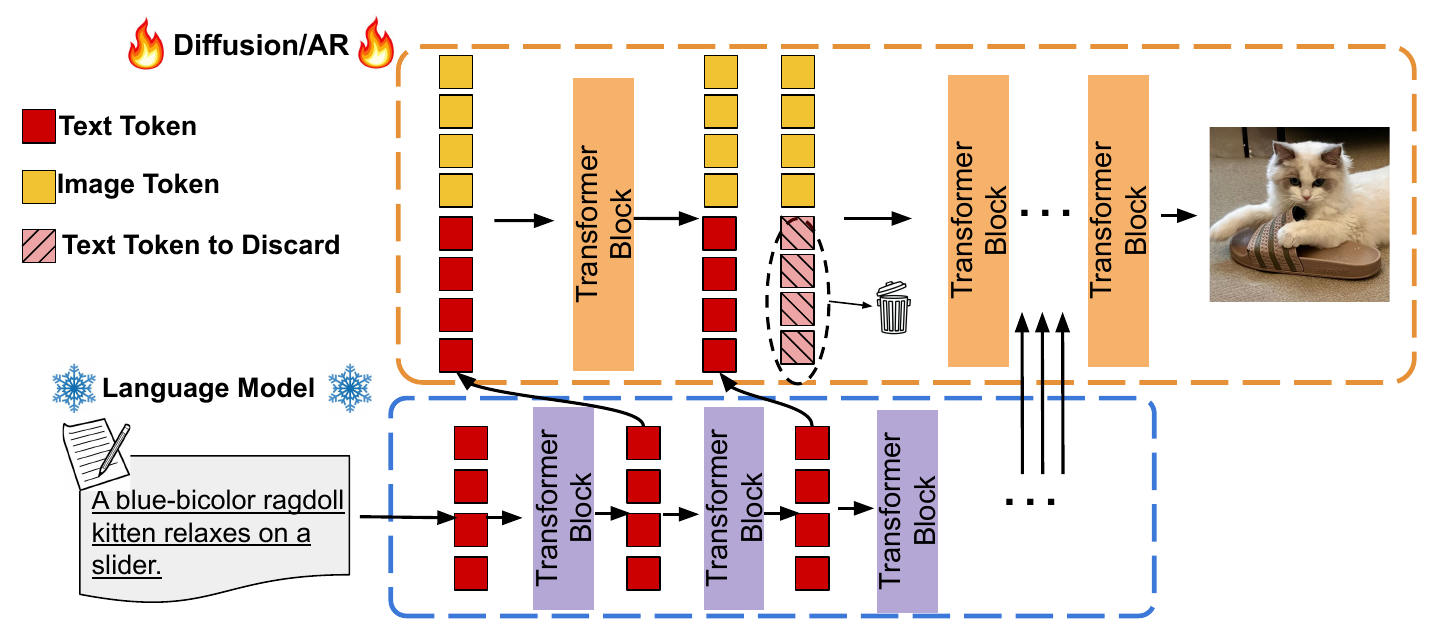}
    \caption{Model architecture of text-to-image diffusion or AR model in 1D latent space. We condition the generative models with intermediate text tokens from a pre-trained language model and fuse with image tokens via joint self-attention layers.}
\label{fig:t2i}
\end{figure}

\paragraph{Model Architectures.} 
Previous text-to-image (T2I) models mainly rely on frozen CLIP \cite{radford2021learning}, T5 \cite{t5}, or a combination of both for text feature extraction. 
In this paper, we build upon the rapid advancements in large language models (LLMs) and investigate the use of decoder-only models as text encoders in T2I models as illustrated in Figure~\ref{fig:t2i}. 
Specifically, our T2I models adopt a dual-stream architecture, where in the text encoding stream, text features are extracted from all Transformer blocks of an LLM, which remain frozen throughout the entire T2I model training. In the image generation stream, we adopt a fully Transformer model with an architecture mirroring that of the LLM. Each Transformer block in the image generation stream receives image features from the preceding block and text features from the same depth of the LLM as input. To align the modalities, a linear layer transforms the text features before they are concatenated with the image features as input to a joint text-image self-attention layer. 
For the diffusion and autoregressive models, we employ bidirectional attention and causal attention for the image tokens, respectively. Since the latent tokens have no predefined spatial order, we use learnable position embeddings initialized randomly in all experiments. Each transformer block learns its own positional embedding, which is added to the image tokens at the start of the block.

In pursuit of fully reproducible foundational model research, we use the fully open-source AMD OLMo-1B\footnote{\href{https://huggingface.co/amd/AMD-OLMo-1B}{huggingface.co/amd/AMD-OLMo-1B}} model as the text feature encoder in our models.

\paragraph{Implementation Details. }
For the text-to-image experiments, we train the 1D binary image tokenizer to encode images into 128 discrete latent tokens, each of which has 64 binary values. We train the binary image tokenizer to reconstruct images at various image resolutions including $512\times512$, $768\times768$, and $1024\times1024$.
We train the tokenizer with a batch size of 1024 for 50K steps and 10K steps for stage-1 and stage-2, respectively. And tokenizer is then trained for another 200K steps with a batch size of 800 for stage-3. 
We use a subset of LAION with $\sim$600M images for the training of all tokenizers, and all trainings are conducted on a single node with 8 MI300X GPUs.
The text-to-image generation models are trained using open data sources, including filtered and recaptioned open data and synthetic data. Details on training data can be found in Appendix Section~\ref{training_data}.
During training, we exclude images for which the shorter edge is less than half the target resolution. Due to time and resource constraints, we perform pretraining only and do not perform extensive filtering or curation. We leave post-training and more comprehensive data curation to future research.
More details can be found in Appendix~\ref{details}.

For both diffusion and auto-regressive models, the model architectures mirror the OLMo-1B, with 16 Transformer blocks, a hidden dimension of 2048, and a total parameter size of 1B. In each Transformer block, a join self-attention with both the text tokens and image tokens is performed with 16 attention heads. Specifically, we exclude text tokens from the query since each Transformer block receives the text tokens from the LLM at the corresponding depth as the text condition, and we do not update the text tokens across the image branch of our generative models. 
Thanks to the reduced token numbers, for all experiments, we can fit 512 training samples into a single AMD MI300X GPU with BF16 data precision and gradient checkpointing, resulting in a global batch size of 4096 on a single 8-GPU computation node.

\paragraph{Results.}
We perform quantitative evaluations on our text-to-image diffusion and auto-regressive image generation models including Stable Diffusion model family \cite{rombach2022highresolutionimagesynthesislatent,podell2023sdxl,esser2024scaling}, PixArt model familty \cite{chen2023pixart,chen2024pixart}, Chameleon \cite{team2024chameleon}, and Emu3\cite{wang2024emu3}. 
In Table~\ref{tab:geneval}, we show results evaluated using multiple text-to-image benchmarks. 
We report results on GenEval \cite{ghosh2024geneval} to assess the compositionality of the generated results, and CLIP \cite{hessel2021clipscore} and ImageReward \cite{xu2023imagereward} scores to evaluate text-image alignment and quality. Despite the fact that most of the state-of-the-art models use their private in-house data for training, our diffusion models, trained on publically available data sources with no post-training, achieve competitive performance with a similar model size. As we will further show in Table~\ref{tab:speed}, our diffusion models with 1B model size demonstrate fast training and sampling speed, with 20-step sampling completed in 0.4 seconds, further closing the speed gap compared to one-step GAN methods even without timestep distillation.
Our fully auto-regressive model with the proposed 1D discrete binary latent space achieves performance comparable to powerful diffusion models. Although our auto-regressive model achieves lower performance compared to Emu3 \cite{wang2024emu3}, it is worth noting that Emu3 is much larger in size and requires 16$\times$ more tokens than our method for image generation. Additionally, Emu3 takes approximately 5 minutes to generate a single image on an MI300X GPU, whereas our model completes generation in only about 3 seconds without leveraging KV caching and $\sim$0.7 seconds with KV caching. Text-to-image generation with faster parallel token decoding AR methods will be explored in the future for further speedup. 
By reducing the number of discrete tokens to 128, our method uses 16–32 times fewer tokens compared to standard VQ-VAE-based auto-regressive image generation approaches. The substantial token reduction and remarkable compatibility with auto-regressive generation highlight the potential of integrating 1D binary latent-space auto-regressive image generation into modern LLMs, enabling joint image-language generation and understanding at remarkable generation speeds.
We present qualitative results in Figure~\ref{fig:teaser} and side-by-side comparisons with images generated by other models in Appendix Section~\ref{sbs}.

Note that rather than aiming to develop the best-performing text-to-image model, our primary goal is to develop highly efficient models that significantly reduce the number of discrete latent tokens required for image generation. Additionally, we curate training recipes using open datasets to ensure easy reproducibility. As detailed further in Section~\ref{efficiency}, every model we report here is trained within 200 GPU-days from scratch on a single 8-GPU computation node, significantly reducing the training budget required to achieve competitive performance evaluated using challenging benchmarks. This approach potentially lowers the barrier to entry, making foundational text-to-image model training accessible to a broader audience, especially those with limited computational resources.

\begin{table*}[t!]
    \centering
    \small
    \caption{Quantitative evaluation results on text-to-image generation. `IR' stands for ImageReward. Details regarding the evaluation protocols are detailed in Appendix Section~\ref{eval}. Our results are indicated in \colorbox{lightgray!20}{Grey}.
    Our AR and binary diffusion models perform comparably on large-resolution images with fewer discrete tokens.
    }
    \label{tab:geneval}
    \resizebox{\linewidth}{!}{
    \begin{tabular}{c|c | c ||  ccc  ccc  | c|| c || c}
    \toprule
    Model & Size & Reso. & Single Obj. & Two Obj. & Counting & Colors & Color Attr. & Position & Overall $\uparrow$ & CLIP $\uparrow$ & IR $\uparrow$\\
    \midrule
    SDv1.5  & 0.9B & 512 & 0.97 & 0.38 & 0.35 & 0.76 & 0.06 & 0.04 & 0.43 & 0.318 & 0.201\\
    SDv2.1  & 0.9B & 512& 0.98 & 0.51 & 0.44 & 0.85 & 0.17 & 0.07 & 0.50 & 0.338 & 0.372 \\
    PixArt-$\alpha$ & 0.6B & 1024 & 0.98 & 0.50 & 0.44 & 0.80 & 0.07 & 0.08 & 0.48 & 0.321 & 0.871 \\
    PixArt-$\sigma$  & 0.6B & 1024 & 0.98 & 0.59 & 0.50 & 0.80 & 0.15 & 0.10 & 0.52 & 0.325 & 0.872\\
    SDXL & 2.6B & 1024 &  0.98 & 0.74 & 0.39 & 0.85 & 0.23 & 0.15 & 0.55 & 0.335 & 0.600\\
    SD3-Medium & 8.0B & 1024 & 0.97 & 0.89 & 0.69 & 0.82 & 0.47 & 0.34 & 0.69 & 0.334 & 0.871 \\
    \midrule
    Chameleon & 7.0B & 512 & - & - & - & - & -  & -& 0.39 & - & -\\
    Emu3 & 8.0B & 1024 & 0.98 & 0.71 & 0.34 & 0.81 & 0.17 & 0.21 & 0.54 & 0.333 & 0.872 \\
    \midrule
    \grayrow 
    Instella AR & 0.8B & 1024 &0.96 & 0.43 & 0.40 & 0.80 & 0.14 & 0.08 & 0.46 & 0.313 & 0.538\\
    \grayrow
    Instella Diff & 1.2B & 1024 & 0.99 & 0.78 & 0.66 & 0.85 & 0.45 & 0.12 & 0.64 & 0.332 & 0.900\\
    \bottomrule
    \end{tabular}
    }
\end{table*}

\subsection{Efficiency}
\label{efficiency}
We present performance and efficiency comparisons in Table~\ref{tab:speed}. 
The significantly reduced number of latent tokens leads to substantial improvements in both training and inference speed, making it possible to complete text-to-image pretraining within 200 GPU-days.
Notebaly, thanks to the multi-resolution decoding capability of our 1D binary image tokenizer, images at different resolutions can now be generated using the same latent, achieving a nearly consistent inference speed despite image resolution. On the opposite, 2D image tokenizers require 4$\times$ more latent tokens with scaling up image resolution by 2$\times$.
It is worth noting that the current training efficiency is achieved without extensive data curation. As suggested in previous works such as \cite{chen2023pixart}, carefully curated training data could reduce the overall volume of data needed, further accelerating the training process. We plan to explore data curation strategies for text-to-image model training in future research. 

\clearpage
\subsection{Reconstruction}
\begin{wrapfigure}{r}{0.6\textwidth}
\vspace{-8mm}
  \centering
    \small
    \captionof{table}{Comparison of various tokenizers on 512$\times$512 ImageNet and MS-COCO. $\dagger$ indicates training on other data than ImageNet. 
    BIT achieves comparable performance with fewer binary tokens.}
    \label{tab:tok_comp}
    \resizebox{\linewidth}{!}{%
    \begin{tabular}{l c c c c c c c c c}
    \toprule
    \multirow{2}{*}{Tokenizer} & \multirow{2}{*}{Size}   & \multirow{2}{*}{\# Tkn}
      & \multicolumn{3}{c}{ImageNet}
      & \multicolumn{3}{c}{COCO} \\
    \cmidrule(lr){4-6}\cmidrule(lr){7-9}
      &  &  & rFID$\downarrow$ & PSNR$\uparrow$ & SSIM$\uparrow$
            & rFID$\downarrow$ & PSNR$\uparrow$ & SSIM$\uparrow$ \\
    \midrule
    \multicolumn{9}{l}{\textit{continuous}}\vspace{0.02in} \\
    \pz\pz SD-VAE$^\dagger$    & 84M  & 4096 &  0.19 & 27.36 & 0.849 & 2.41  & 26.48 & 0.841  \\
    \pz\pz DC-AE$^\dagger$  &  323M  & 256 & 0.21  &  26.23 & 0.815 & 2.85 & 25.47  & 0.811  \\

    \pz\pz TexTok      &  176M & 256 &  0.73 & 24.45 & 0.668 & - & -  & -  \\

    \pz\pz MAETok       & 176M & 128 & 0.62  & 22.18  & 0.701 & 5.91 & 22.48 & 0.695 \\
    \arrayrulecolor{gray}\cmidrule(lr){1-9}

    \multicolumn{9}{l}{\textit{discrete}}\vspace{0.02in} \\ 

    \pz\pz VQ-GAN       & 66M & 1024 &  5.37 &  20.51 & 0.631  & 9.48 & 19.54 & 0.618 \\
    \pz\pz TiTok-B       & 202M & 128 & 1.52  & -  & -  & - & - & - \\
    \pz\pz LlamaGen       & 116M & 1024 & 0.87  &  - & -  & - & - & - \\
    
    \grayrow
    \pz\pz BIT       & 397M & 128 & 1.32  & 22.25  &  0.704 & 7.14 & 21.59 & 0.69 \\

    \bottomrule
    \end{tabular}
    }
    \\
    \captionof{table}{Efficiency comparisons against other text-to-image models. The `Steps' column refers to the number of inference steps for the generation of each sample.}
    \label{tab:speed}
    \resizebox{\linewidth}{!}{
    \begin{tabular}{c|c c c c c}
    \toprule
    Model & Size & Steps & FID-30K $\downarrow$ & Sec/img $\downarrow$ & GPU-days \\
    \midrule
    SDv1.5 & 0.9B & 50 & 14.35 & 1.34 &  6,250 A100\\
    PixArt-$\alpha$ & 0.6B & 20 & 21.30 & 2.28 & 753 A100\\
    \grayrow
    Diffusion BIT & 1B & 20  & 16.33 & 0.38 & 177 MI300X \\
    \grayrow
    Diffusion BIT & 1B & 50  & 15.10 & 1.32 & 177 MI300X \\
    \bottomrule
    \end{tabular}
    }
\vspace{-5mm}
\end{wrapfigure}

The quality of image reconstruction is a crucial factor in evaluating tokenizers.
As shown in Table~\ref{tab:tok_comp}, our proposed 1D binary latent tokenizer (BIT) achieves competitive reconstruction quality while significantly reducing the number of tokens compared to traditional continuous and discrete tokenizers.
Notably, BIT with just 128 tokens maintains a reasonable trade-off between reconstruction fidelity and efficiency, with an rFID score of 1.32 and SSIM of 0.704 on ImageNet 512$\times$512 reconstruction.
While SD-VAE achieves superior rFID and SSIM scores, it requires 4096 continuous tokens over 30 times more than BIT’s 128-token configuration. Moreover, BIT maintains a consistent token number of 128 when further scaling up the image resolution to 1024$\times$1024. The ability to achieve competitive results with such a drastic reduction in token count underscores the efficiency and scalability of our method.

\section{Conclusion}
We introduce efficient image generation in 1D binary latent space. Using Transformer-based tokenizers, our approach encodes a 1024$\times$1024 image into as few as 128 discrete tokens, each represented by a binary vector. This compact latent representation supports diverse generative models, including diffusion and auto-regressive methods, enabling both high training and inference speeds. Through diffusion and auto-regressive text-to-image model training, we demonstrate that the proposed 1D binary latent achieves competitive generation quality, offering a scalable and efficient solution to modern image generation challenges. Further directions for research are discussed in Appendix Section~\ref{future}.


\bibliography{main}

\clearpage
\setcounter{page}{1}
\appendix

\setcounter{table}{0}
\setcounter{equation}{0}
\setcounter{figure}{0}

\renewcommand{\thetable}{\Alph{table}}
\renewcommand{\thefigure}{\Alph{figure}}
\onecolumn

\section{Text-to-Image Prompts}
\label{prompts}

The prompts, model, and inference configurations used for the text-to-image samples demonstrated in Figure~\ref{fig:teaser} are listed below:

{\small
\begin{itemize}
    \item \textit{A fashion style pretty woman portrait wearing luxurious detailed colorful flower earrings made from leather scraps in the style of liam sharp, dutch tradition, webcam, gail simone, mark seliger, Alfons Mucha inspired 8k, high resolution, dark background.} Diffusion 1024$\times$1024, $\alpha=9.0$ $\tau=0.8$, 20 steps. 
    
    \item \textit{Medium shot, adorable creature with big reflective eyes, moody lighting, best quality, full body portrait, real picture, intricate details, depth of field, in a forest, fujifilm xt3, outdoors, bright day, beautiful lighting, raw photo, 8k uhd, film grain, unreal engine 5, ray tracing.} Diffusion 1024$\times$1024, $\alpha=7.5$ $\tau=1.0$, 50 steps.

    \item \textit{Cute robot in a massive crowded scifi city at night, sketch, rough, first draft, intricate details, cute expression, big eyes, Minimalist color sketch of sci-fi.} Diffusion 512$\times$512, $\alpha=7.5$ $\tau=0.75$, 20 steps. 

    \item \textit{The ethereal beauty of a mystical landscape under the red moonlight. The scene should be illuminated by a large, radiant moon, casting its glow upon a twisted, yet majestic tree with blossoms that seem to sparkle in the night. The tree’s roots should be deeply embedded into rocky terrain, symbolizing its ancient existence. In the background, towering mountains loom, their peaks veiled in mist. A serene lake at the foot of the mountains reflects the moon’s luminescence. Incorporate elements that evoke a sense of magic and mystery} Diffusion 512$\times$512, $\alpha=9.0$ $\tau=0.8$, 20 steps. 
    
    \item \textit{A hummingbird, Hyperdetailed Eyes, Tee-Shirt Design, Line Art, Black Background, Ultra Detailed Artistic, Detailed Gorgeous Face, Natural Skin, Water Splash, Colour Splash Art, Fire and Ice, Splatter, Black Ink, Liquid Melting, Dreamy, Glowing, Glamour, Glimmer, Shadows, Oil On Canvas, Brush Strokes, Smooth, Ultra High Definition, 8k, Unreal Engine 5, Ultra Sharp Focus, Intricate Artwork Masterpiece, Ominous, Golden Ratio, Highly Detailed, Vibrant, Production Cinematic Character Render, Ultra High Quality Model.} Diffusion 512$\times$512, $\alpha=7.5$ $\tau=0.75$, 20 steps. 

    \item \textit{Pirate ship sailing into a bioluminescence sea with a galaxy in the sky), epic, 4k, ultra.} Diffusion 512$\times$512, $\alpha=6.0$ $\tau=0.75$, 100 steps. 
    
    \item \textit{Abstract oil painting of Chrsimple 3D front view icon made out of polygons of a blue, green, yellow, orange, red, purple, and white futuristic sneaker, on a white backgroundistmas tree, pastel colors splash.} Diffusion 512$\times$512, $\alpha=7.5$ $\tau=0.75$, 20 steps. 

    \item \textit{A rainforest suspended in a crystal sphere floating in space, with clouds swirling around its equator.} Diffusion 512$\times$512  $\alpha=7.5$ $\tau=1.0$, 20 steps. 

    \item \textit{A simple 3D front view icon made out of polygons of a blue, green, yellow, orange, red, purple, and white futuristic sneaker, on a white background.} Diffusion 512$\times$512, $\alpha=8.0$ $\tau=0.9$.
    \item \textit{Morning light streams through the window of a cluttered artist's studio, illuminating open sketchbooks and half-finished paintings. } Diffusion 512$\times$512  $\alpha=7.5$ $\tau=1.0$, 50 steps. 
    
    \item \textit{Watercolor paint of new yor city, colored, vivid, bright.} Auto-regressiv 512$\times$512,$\alpha=7.5$ $\tau=0.6$, 20 steps. 
    
    \item \textit{Detailed pen and ink illustration of a samurai robot cyborg, head only, detailed anthropology, by Herge, in the style of tin-tin comics, detailed, high quality.} Auto-regressive 512$\times$512,$\alpha=9.0$ $\tau=0.6$.

\end{itemize}
}

\section{Additional Training Details}
\label{details}
If not otherwise specified, we use a single computation node with 8 AMD MI300 GPUs for the training of each experiments. 
For text-to-image diffusion and auto-regressive models, DeepSpeed ZeRO-2, gradient checkpointing, and BF16 mix precision training are used to enable a global batch size of 4096 on a single computation node for all text-to-image experiments. For all tokenizer and generative model training, we use the AdamW optimizer with a consistent learning rate of 1e-4 and 1K steps of learning rate warmup. 
We do not perform any model sharding to the tokenizer training due to the relatively small model sizes. 
We use gradient clipping with a maximum gradient scale of 1.0 for all tokenizer and generative model training. For all the text-to-image experiments, we set the maximum sequence length for the text prompt tokenization to 128, which we observe is sufficient to accommodate most of the text prompts. 

\subsection{Training Data}
\label{training_data}
We use LAION-COCO\footnote{\href{https://huggingface.co/datasets/laion/laion-coco}{huggingface.co/datasets/laion/laion-coco}} which comprises over 600M natural images. We additionally annotate the images with aesthetic scores, OCR density, and OCR size, and remove images with low aesthetic scores or dense and small text. The dataset size reduces to 100M after filtering. We recaption the images using the Qwen2.5-VL-7B model \footnote{\href{https://huggingface.co/Qwen/Qwen2.5-VL-7B-Instruct}{huggingface.co/Qwen/Qwen2.5-VL-7B-Instruct}} and generate one long caption and one short caption for each image.
In addtion to LAION, we create a synthetic datasets using FLUX.1-schnell \cite{flux2024}, PixArt-Sigma \cite{chen2024pixart}, and stable diffusion 3.5 \cite{esser2024scaling}. We use the prompts from DiffusionDB\cite{wang2022diffusiondb} and generate one images for each prompt and model. 
We also include data from Dalle-1M\footnote{\href{https://huggingface.co/datasets/ProGamerGov/synthetic-dataset-1m-dalle3-high-quality-captions}{huggingface.co/datasets/ProGamerGov/synthetic-dataset-1m-dalle3-high-quality-captions}} for training.

\section{Sampling Algorithms}
\label{algo}

For the diffusion inference, we have two sampling algorithms, including the original sampling algorithm Alg.~\ref{alg:sampling} as in \cite{bnl}, and a simplified sampling algorithm Alg.~\ref{alg:sampling2} that we observe is particularly suitable for very low sampling steps.

\begin{algorithm}[]
	\small
	\caption{Orignal sampling algorithm as described in \cite{bnl}.}
	\label{alg:sampling}
	\begin{algorithmic}[1]
		\STATE {\textbf{Given}: Trained decoder $\dec$; Trained binary diffusion model $f_\theta$.} 
		\STATE{Temperature $\tau$; Number of sampling steps $S$; Latent dimension specified as $k \times \hat{c}$. }
        \STATE{Sampling $\latent^1 = \mathcal{B}(0.5 \cdot \mathbf{1}_{k \times \hat{c}})$, }
        \FOR{Step $s = 1:S$}
        \STATE{Current timestep: $tm1 = 1-(s-1)/S$; Next time step $t = 1-s/S$}
        \STATE{Predicting $p_\theta(\latent^{t})$ with $p_\theta(\latent, t) = \sigma(f_\theta(\latent^{tm1}, tm1) / \tau)$ and Eq. 8 of \cite{bnl}.}
        \STATE{Sampling $\latent^{t} = \mathcal{B}(p_\theta(\latent^{t}))$}
		\ENDFOR
		\STATE{\textbf{Return} the sampled image as $\dec(\latent^{0})$.}
	\end{algorithmic}
\end{algorithm}

\begin{algorithm}[]
	\small
	\caption{Samplified sampling algorithm for diffusion in binary latent space.}
	\label{alg:sampling2}
	\begin{algorithmic}[1]
		\STATE {\textbf{Given}: Trained decoder $\dec$; Trained binary diffusion model $f_\theta$.} 
		\STATE{Temperature $\tau$; Number of sampling steps $S$; Latent dimension specified as $k \times \hat{c}$. }
        \STATE{Sampling $\latent^1 = \mathcal{B}(0.5 \cdot \mathbf{1}_{k \times \hat{c}})$, }
        \FOR{Step $s = 1:S$}
        \STATE{Current timestep: $tm1 = 1-(s-1)/S$; Next time step $t = 1-s/S$}
        \STATE{Predicting $p_\theta(\latent)$ with $p_\theta(\latent, t) = \sigma(f_\theta(\latent^{tm1}, tm1) / \tau)$.}
        \STATE{Adding noise to timestep $t$: $p_\theta(\latent^t) = 0.5t + (1-t)p_\theta(\latent, t)$.}
        \STATE{Sampling $\latent^{t} = \mathcal{B}(p_\theta(\latent^{t}))$}
		\ENDFOR
		\STATE{\textbf{Return} the sampled image as $\dec(\latent^{0})$.}
	\end{algorithmic}
\end{algorithm}

\section{Evaluation Details}
\label{eval}

\subsection{Inference configuration}

The results of our models in Table~\ref{tab:geneval} are measured using sampling Alg.~\ref{alg:sampling} with $\alpha=7.5$, $\tau=0.75$, and 100 sampling steps. 
The results of our models in Table~\ref{tab:speed} are measured using sampling Alg.~\ref{alg:sampling2} with $\alpha=4.0$, $\tau=0.6$, and 20 sampling steps. 
The qualitative results of our models in Section~\ref{sbs} are measured using sampling Alg.~\ref{alg:sampling} with $\alpha=7.5$, $\tau=0.75$, and 50 sampling steps. 
\subsection{Metric details}

Due to the potential mismatches about the detailed evaluation protocols across previous works, we present in the section the details of the evaluation protocol for each metric used in our work.

\paragraph{CLIP score.} We use MSCOCO 2014 to evaluate the CLIP scores reported in Table~\ref{tab:geneval}. For each model, we generate 30K images using the standard prompts. The CLIP score for each sample is evaluated using the pretrained ViT-G/14 model with the image resized to $256\times256$. The final reported number is an average over all the 30K samples. 

\paragraph{FID score.} For text-to-image model evaluation, we adopt the evaluation code of \cite{kang2023scaling}. The images are generated using 3K text prompts from the MSCOCO val2024 set, with one image per prompt. The FID is compared with the MSCOCO val2014 set, and all images are resized to $256\times256$ before feeding to the Inception V3 model. 

\paragraph{ImageReward.} ImageReward scores are calculated using the official evaluation code\footnote{\href{https://github.com/THUDM/ImageReward}{github.com/THUDM/ImageReward}}, with a total of 100 text prompts and one image generated for each prompt. 

\paragraph{GenEval.} We use the official code\footnote{\href{https://github.com/djghosh13/geneval}{github.com/djghosh13/geneval}} for the evaluation of the GenEval scores, with a total of 553 text prompts and 4 images generated for each prompt.

\section{Limitation and Future Work}
\label{future}
\paragraph{Flexible image sizes.} Unlike traditional convolution-based image tokenizers, 1D image tokenizers do not naturally generalize to unseen resolutions and aspect ratios, demonstrating limited flexibility for image sizes. In this work, we train the tokenizer to decode images within a set of resolution, including 512$\times$512, 768$\times$768, and 1024$\times$1024. We leave training more flexible image tokenizers and generative models in 1D binary latent space with compatibility and generalization to variable resolutions and aspect ratios in further efforts. 

\paragraph{Reconstruction articacts.} The current version of the binary 1D image tokenizer can still show noticeable artifacts, especially in regions with high-frequency textures. We believe that the primary reason is the fully transformer model architecture. We notice that reducing the patch size from 16 to 8 can significantly improve the reconstruction quality. However, it introduces nearly prohibitive cost and memory footprints for high-resolution images. We leave potential solutions such as hybrid tokenizer model architectures and diffusion decoders in future efforts.

\paragraph{FID scores.} The current version of text-to-image models does not excel in FID scores. Specifically, Alg~\ref{alg:sampling} demonstrates a high generated image diversity, but can sometimes lead to divergence in the sampling process. Alg~\ref{alg:sampling2} generates images with consistently higher quality but relatively insufficient divesity. We are actively exploring better sampling strategies and image generation in other latent space.

\paragraph{Post training.} The current text-to-image models are pre-trained only, making them incapable of competing more advanced models such as FLux. We will explore preference tuning, guidance and time-step distillation in the proposed 1D binary latent space in future work. 

\paragraph{Unified image-language models.} The compact latent space with drastically reduced token numbers and the promising results shown by the auto-regressive image generation experiments have proven the potential of integrating 1D binary latent space image generation into modern large language models for joint cross-modality understanding and generation with unprecedented image generation speed.

\section{Broader Impacts}

Our image generation model, like other generative models, shares the potential risks of misuse, including the creation of misleading, inappropriate, or harmful content. As with existing systems, careful consideration is needed to mitigate these risks through responsible deployment and use. At the same time, our model introduces significant benefits by enabling fast training with reduced computational costs. This lower barrier to entry broadens accessibility to high-quality generative models, particularly for researchers and practitioners with limited resources. By democratizing access to state-of-the-art text-to-image capabilities, our work has the potential to accelerate innovation, support educational efforts, and foster a more inclusive research community.

\section{Qualitative Results}
\label{sbs}

\begin{figure}[]
    \centering
    \begin{subfigure}[b]{\textwidth}
        \captionsetup{labelformat=empty}
    \begin{subfigure}[b]{0.23\textwidth}
        \captionsetup{labelformat=empty}
        \centering
        \includegraphics[width=\textwidth]{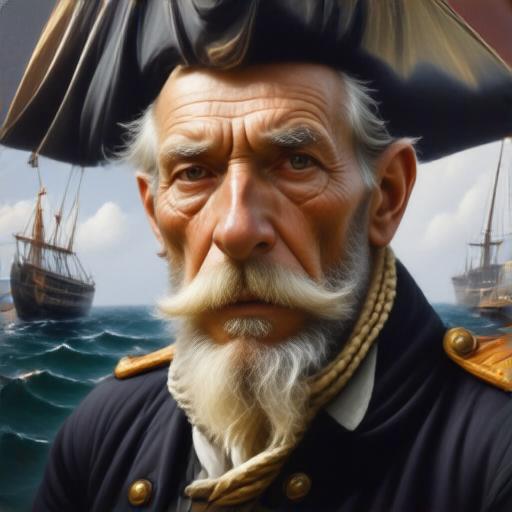}
        \caption{Ours}
    \end{subfigure}
    \hfill
    \begin{subfigure}[b]{0.23\textwidth}
        \captionsetup{labelformat=empty}
        \centering
        \includegraphics[width=\textwidth]{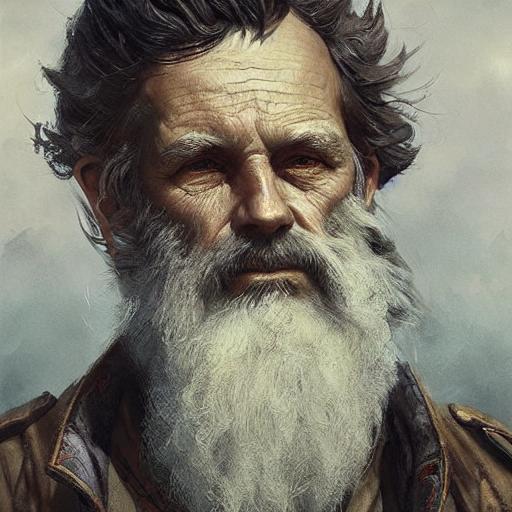}
        \caption{SDv1.5}
    \end{subfigure}
    \hfill
    \begin{subfigure}[b]{0.23\textwidth}
        \captionsetup{labelformat=empty}
        \centering
        \includegraphics[width=\textwidth]{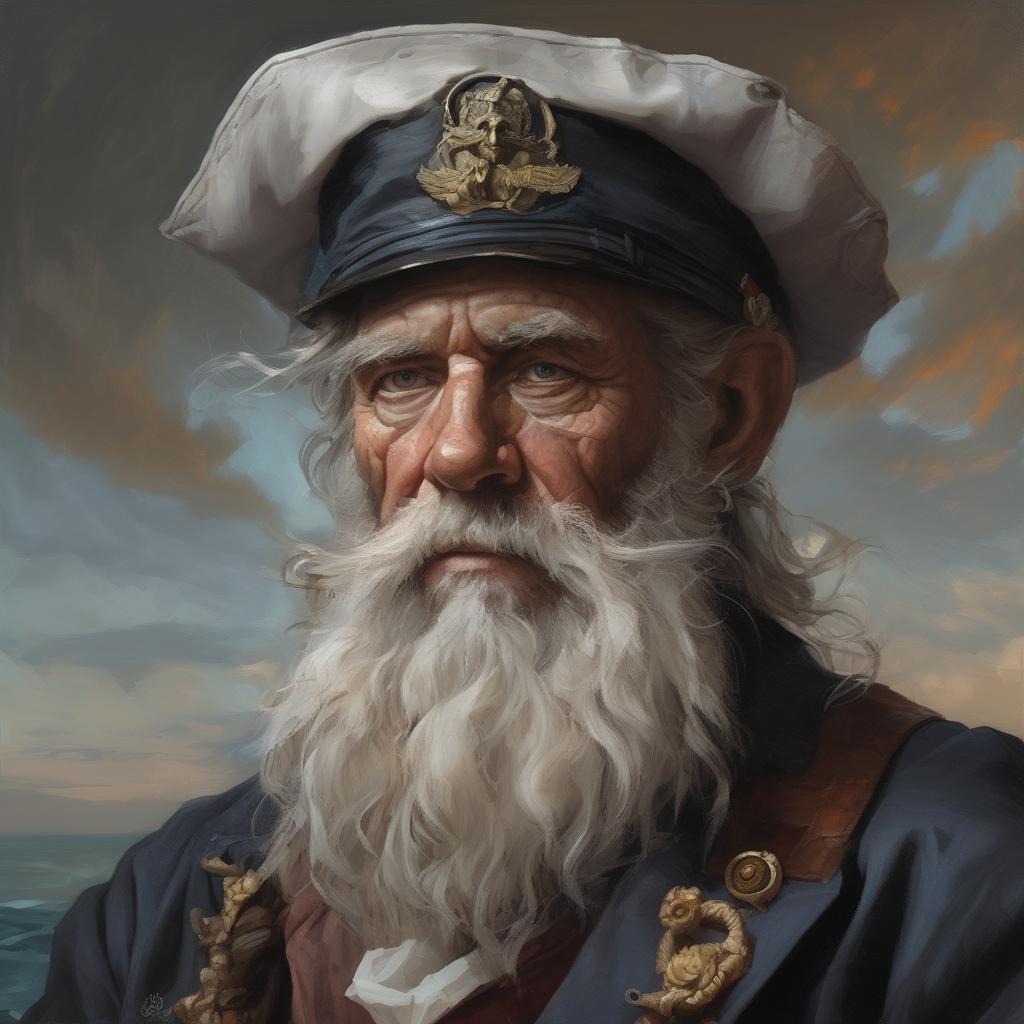}
        \caption{SDXL}
    \end{subfigure}
    \hfill
    \begin{subfigure}[b]{0.23\textwidth}
        \captionsetup{labelformat=empty}
        \centering
        \includegraphics[width=\textwidth]{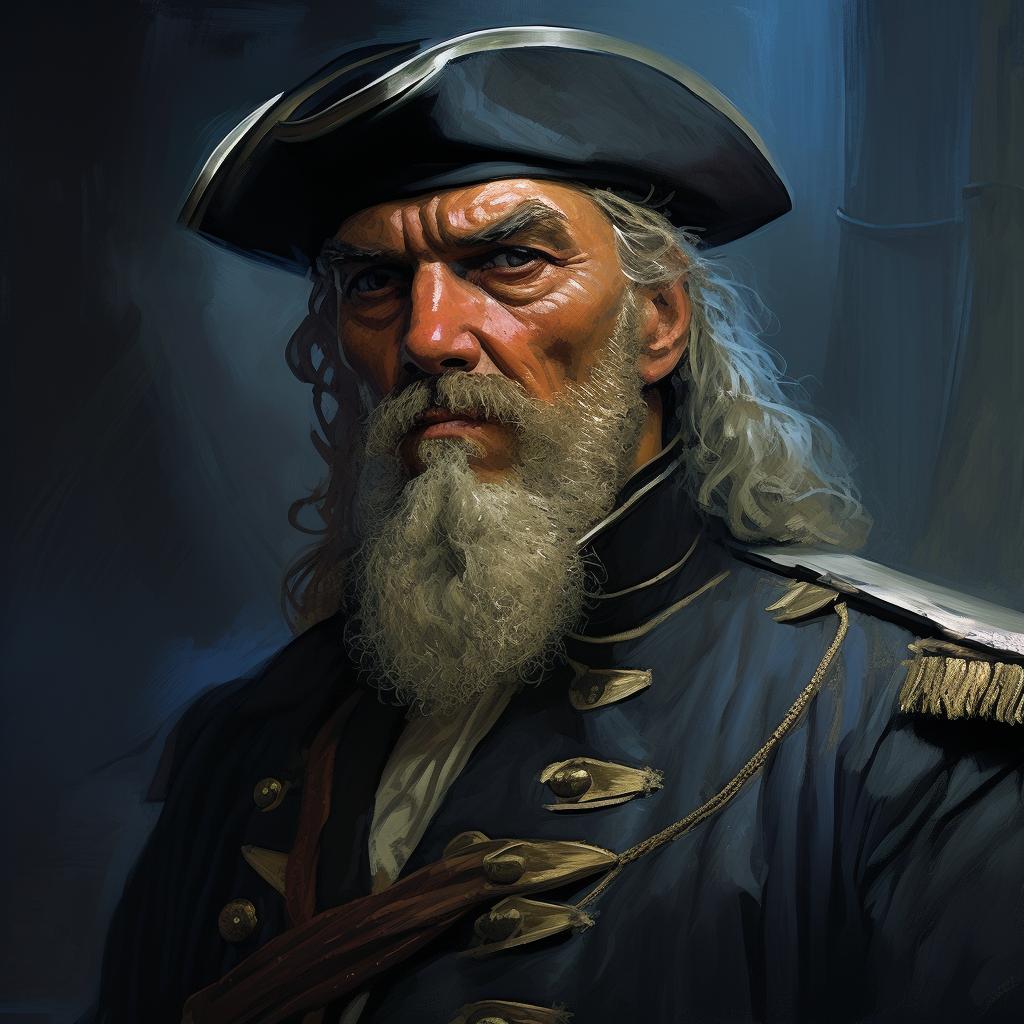}
        \caption{PixArt-Alpha}
    \end{subfigure}
    \captionsetup{labelformat=empty}
    \caption{Portrait of an old sea captain, male, detailed face, fantasy, highly detailed, cinematic, art painting by greg rutkowski.}
    \end{subfigure}

    \vspace{0.5cm} 

    \begin{subfigure}[b]{\textwidth}
        \captionsetup{labelformat=empty}
    \begin{subfigure}[b]{0.23\textwidth}
        \captionsetup{labelformat=empty}
        \centering
        \includegraphics[width=\textwidth]{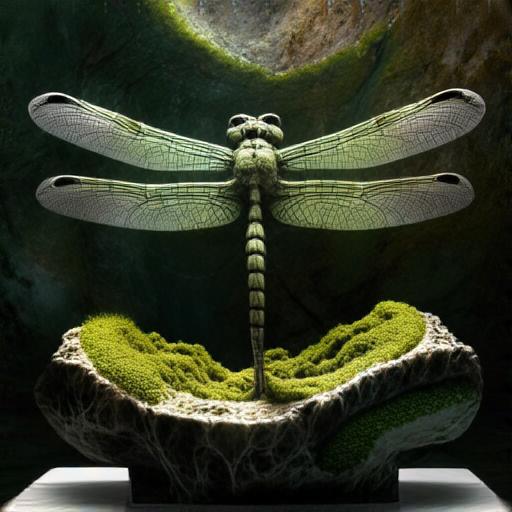}
        \caption{Ours}
    \end{subfigure}
    \hfill
    \begin{subfigure}[b]{0.23\textwidth}
        \captionsetup{labelformat=empty}
        \centering
        \includegraphics[width=\textwidth]{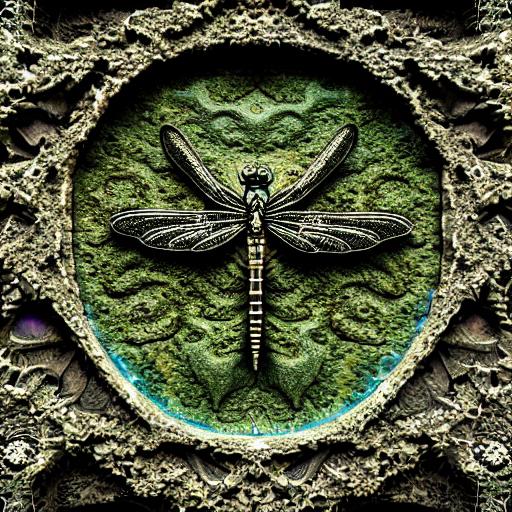}
        \caption{SDv1.5}
    \end{subfigure}
    \hfill
    \begin{subfigure}[b]{0.23\textwidth}
        \captionsetup{labelformat=empty}
        \centering
        \includegraphics[width=\textwidth]{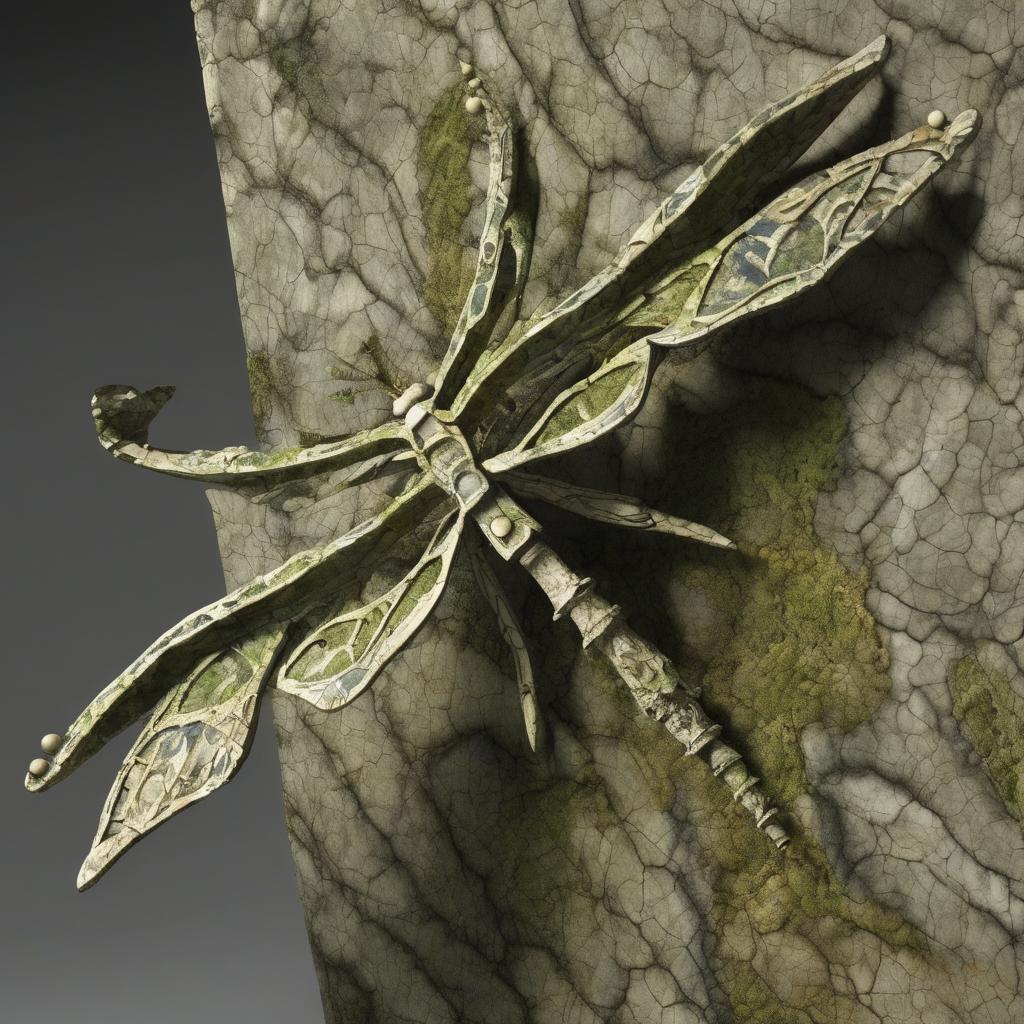}
        \caption{SDXL}
    \end{subfigure}
    \hfill
    \begin{subfigure}[b]{0.23\textwidth}
        \captionsetup{labelformat=empty}
        \centering
        \includegraphics[width=\textwidth]{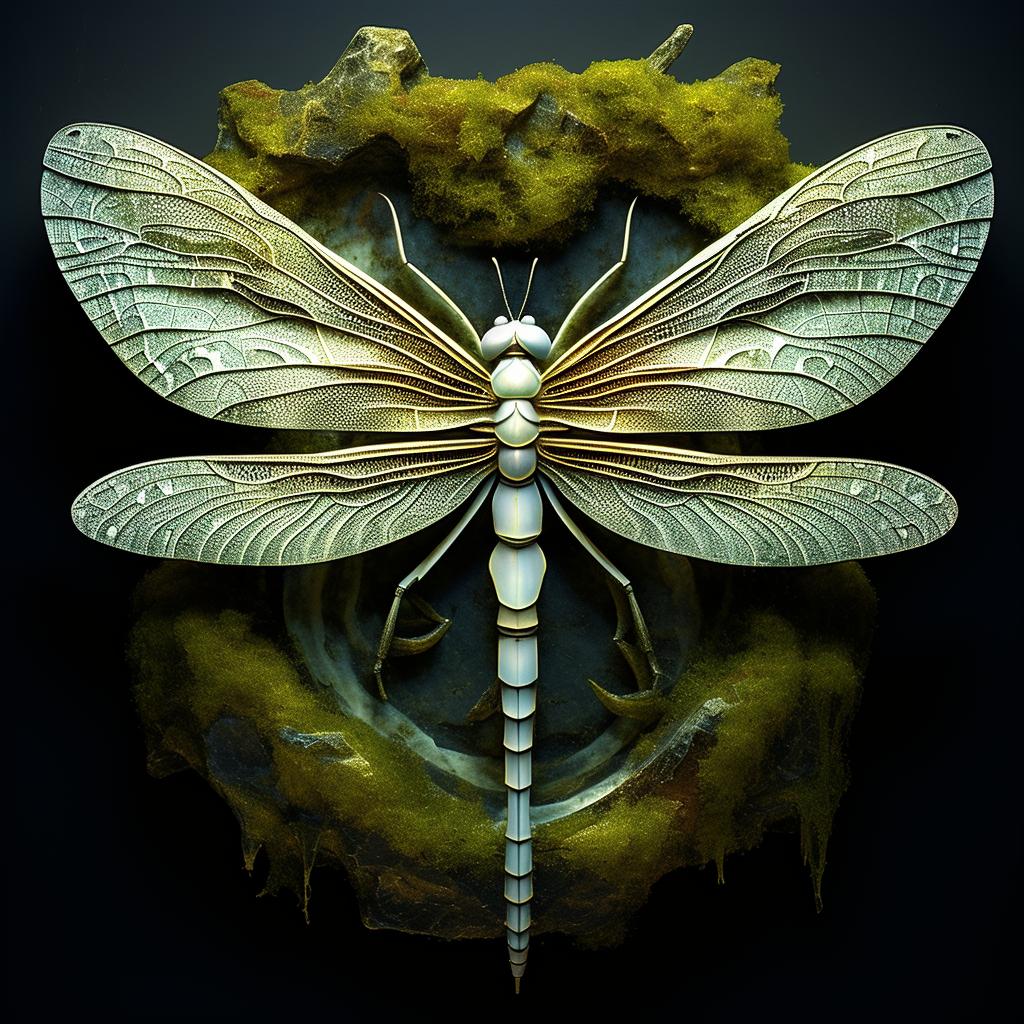}
        \caption{PixArt-Alpha}
    \end{subfigure}
    \captionsetup{labelformat=empty}
    \caption{Dramatically lit, intricate fractal sculpture of an ancient, weathered, mossy marble sculpture of a dragonfly.}
    \end{subfigure}

    \vspace{0.5cm} 

    \begin{subfigure}[b]{\textwidth}
        \captionsetup{labelformat=empty}
    \begin{subfigure}[b]{0.23\textwidth}
        \captionsetup{labelformat=empty}
        \centering
        \includegraphics[width=\textwidth]{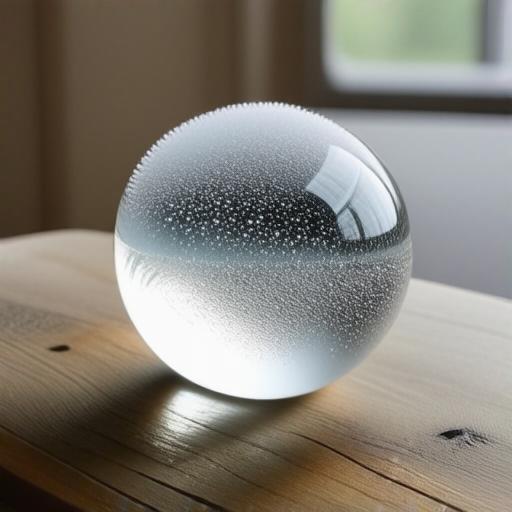}
        \caption{Ours}
    \end{subfigure}
    \hfill
    \begin{subfigure}[b]{0.23\textwidth}
        \captionsetup{labelformat=empty}
        \centering
        \includegraphics[width=\textwidth]{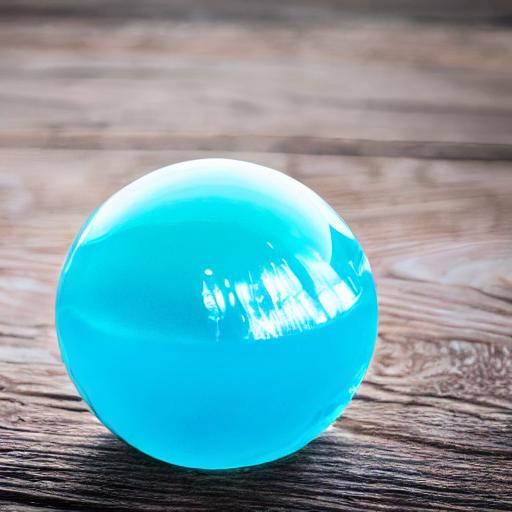}
        \caption{SDv1.5}
    \end{subfigure}
    \hfill
    \begin{subfigure}[b]{0.23\textwidth}
        \captionsetup{labelformat=empty}
        \centering
        \includegraphics[width=\textwidth]{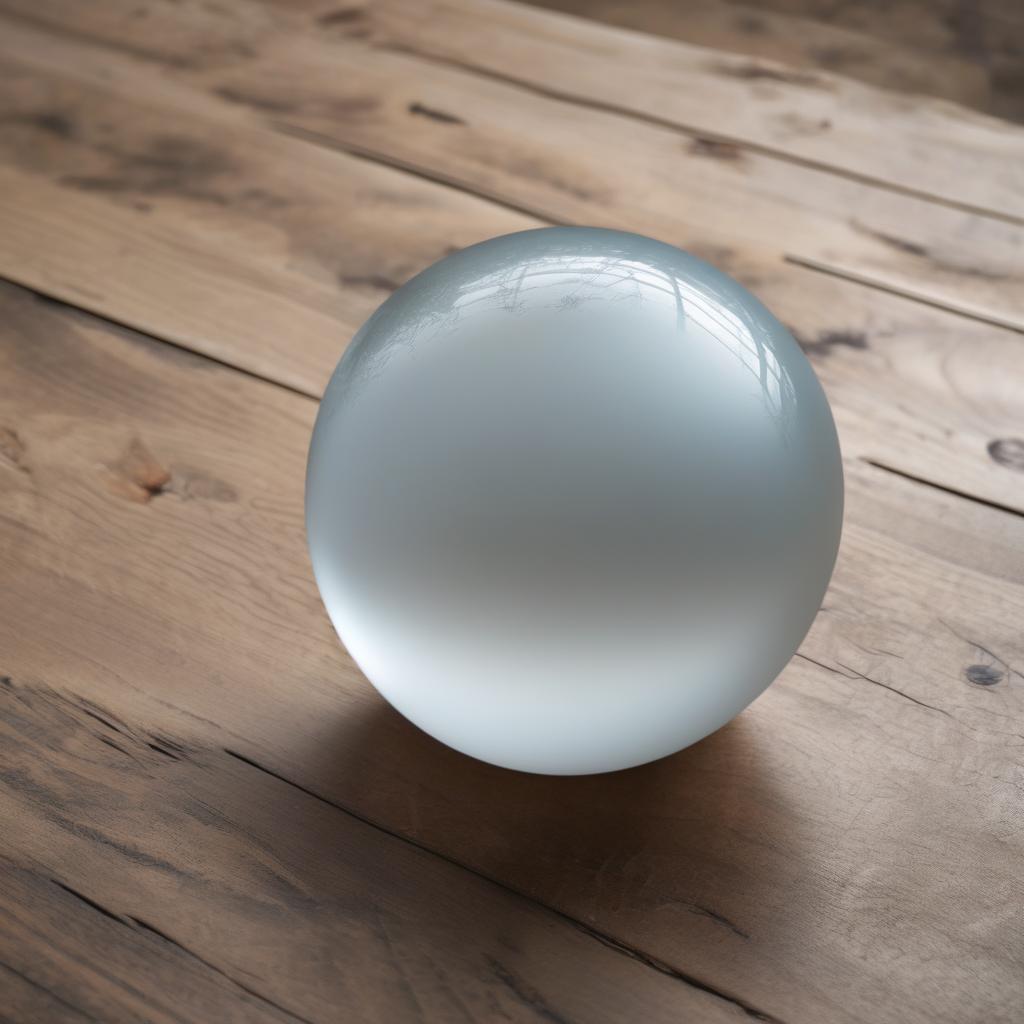}
        \caption{SDXL}
    \end{subfigure}
    \hfill
    \begin{subfigure}[b]{0.23\textwidth}
        \captionsetup{labelformat=empty}
        \centering
        \includegraphics[width=\textwidth]{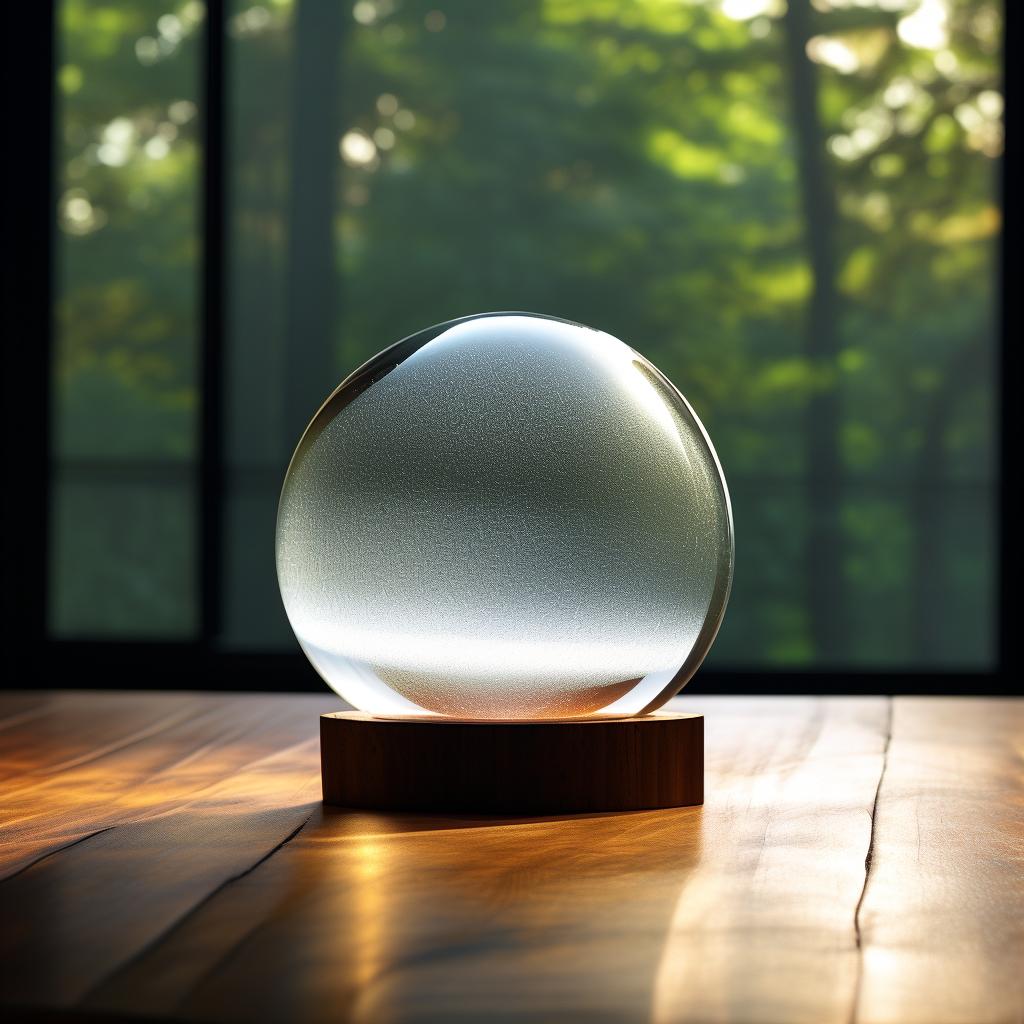}
        \caption{PixArt-Alpha}
    \end{subfigure}
    \captionsetup{labelformat=empty}
    \caption{Frosted glass sphere sitting on a wooden table, high, complex.}
    \end{subfigure}

    \vspace{0.5cm} 

    \begin{subfigure}[b]{\textwidth}
        \captionsetup{labelformat=empty}
    \begin{subfigure}[b]{0.23\textwidth}
        \captionsetup{labelformat=empty}
        \centering
        \includegraphics[width=\textwidth]{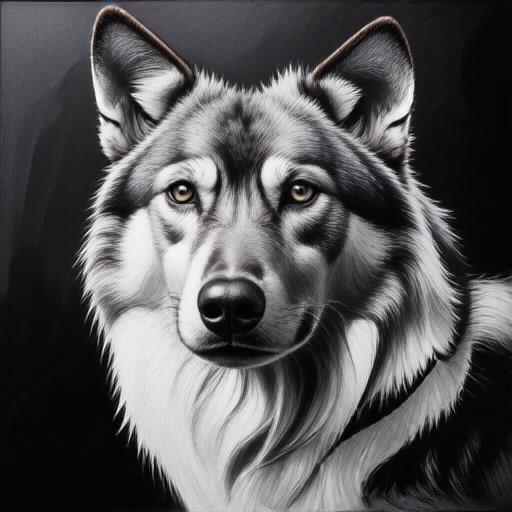}
        \caption{Ours}
    \end{subfigure}
    \hfill
    \begin{subfigure}[b]{0.23\textwidth}
        \captionsetup{labelformat=empty}
        \centering
        \includegraphics[width=\textwidth]{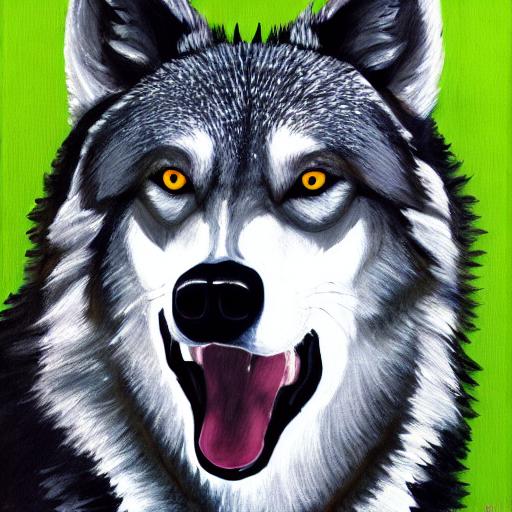}
        \caption{SDv1.5}
    \end{subfigure}
    \hfill
    \begin{subfigure}[b]{0.23\textwidth}
        \captionsetup{labelformat=empty}
        \centering
        \includegraphics[width=\textwidth]{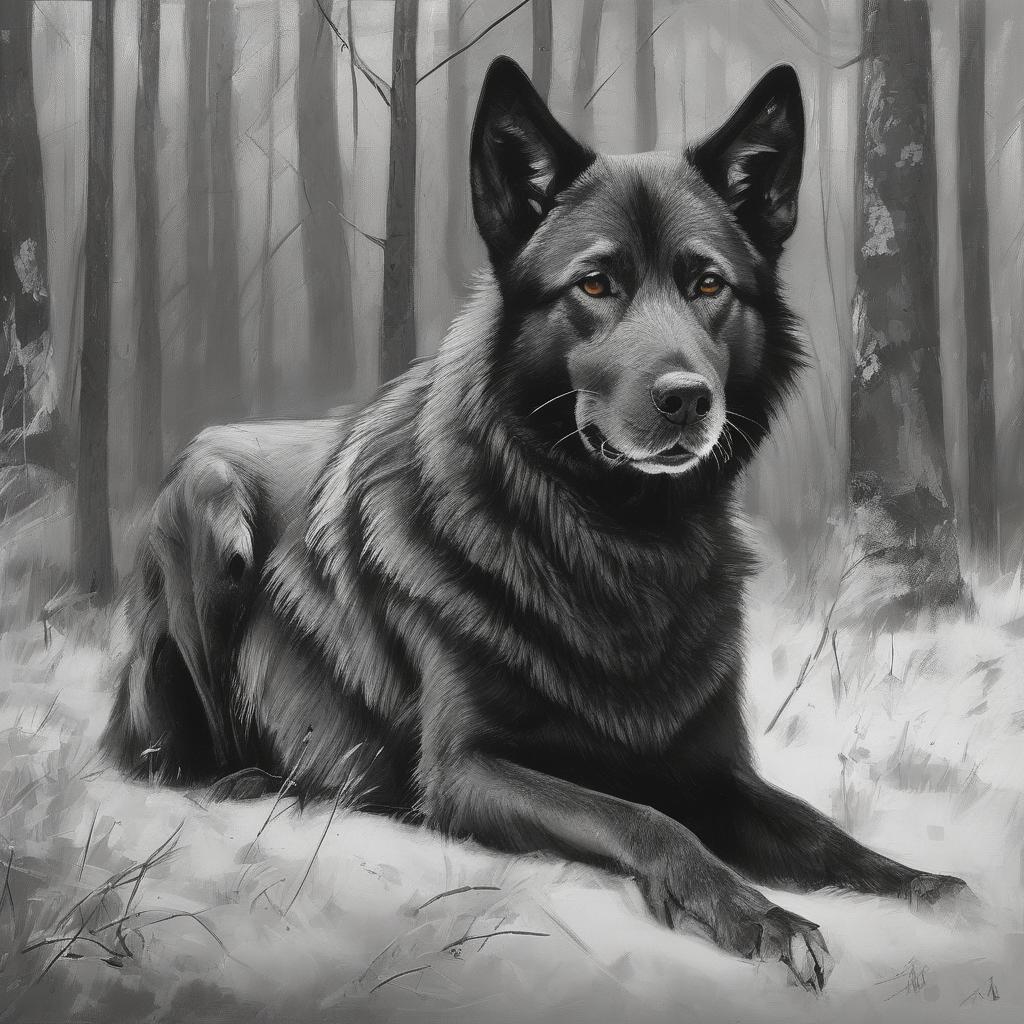}
        \caption{SDXL}
    \end{subfigure}
    \hfill
    \begin{subfigure}[b]{0.23\textwidth}
        \captionsetup{labelformat=empty}
        \centering
        \includegraphics[width=\textwidth]{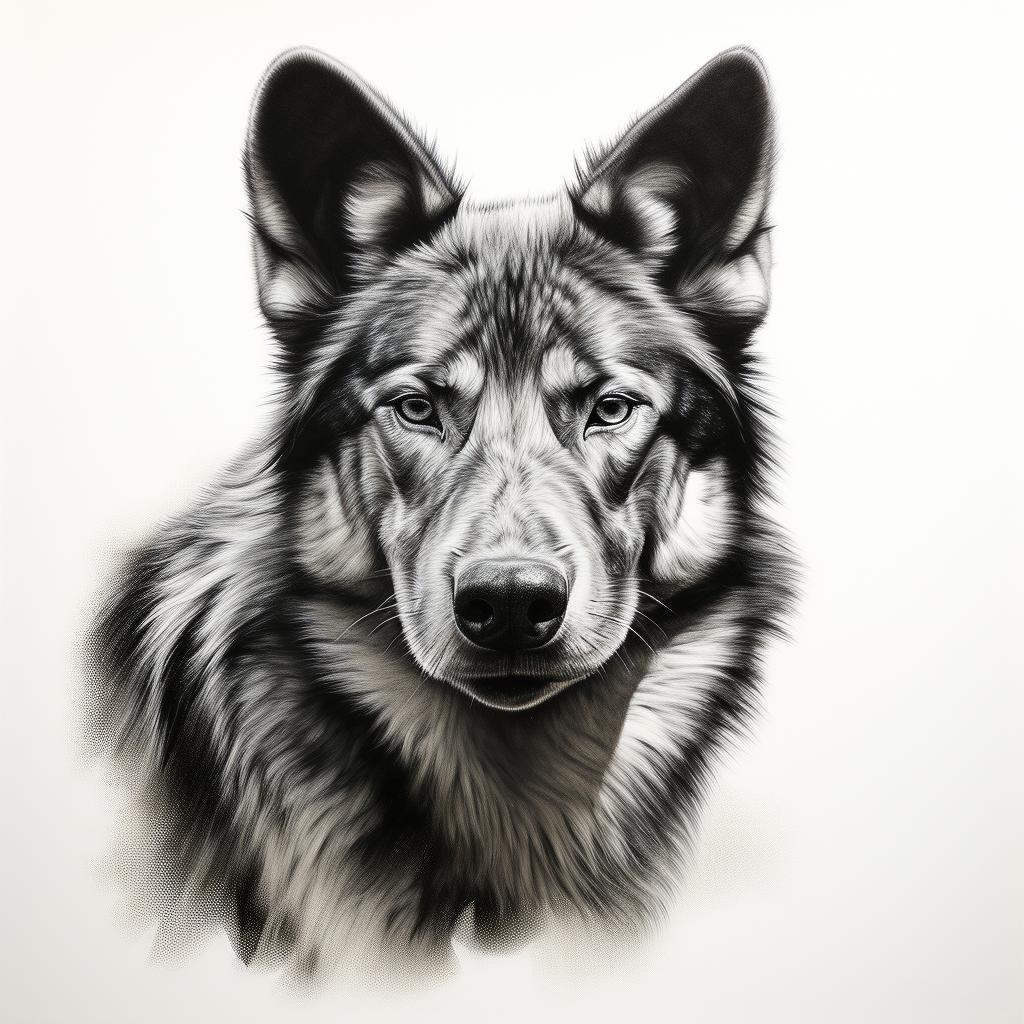}
        \caption{PixArt-Alpha}
    \end{subfigure}
    \captionsetup{labelformat=empty}
    \caption{A checoslovaquian wolfdog, black and white painting.}
    \end{subfigure}
    \caption{Side-by-side comparison of images generated by different methods.}
    \label{fig:sds1}
\end{figure}

\begin{figure}[]
    \centering

    \begin{subfigure}[b]{\textwidth}
        \captionsetup{labelformat=empty}
    \begin{subfigure}[b]{0.23\textwidth}
        \captionsetup{labelformat=empty}
        \centering
        \includegraphics[width=\textwidth]{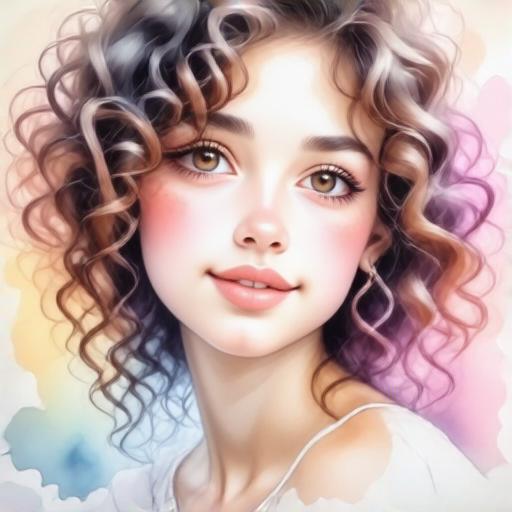}
        \caption{Ours}
    \end{subfigure}
    \hfill
    \begin{subfigure}[b]{0.23\textwidth}
        \captionsetup{labelformat=empty}
        \centering
        \includegraphics[width=\textwidth]{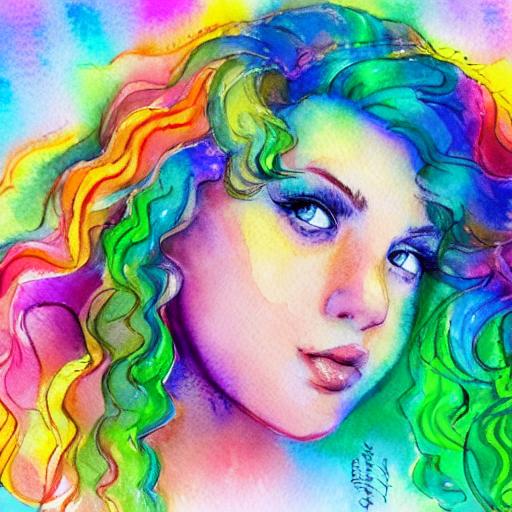}
        \caption{SDv1.5}
    \end{subfigure}
    \hfill
    \begin{subfigure}[b]{0.23\textwidth}
        \captionsetup{labelformat=empty}
        \centering
        \includegraphics[width=\textwidth]{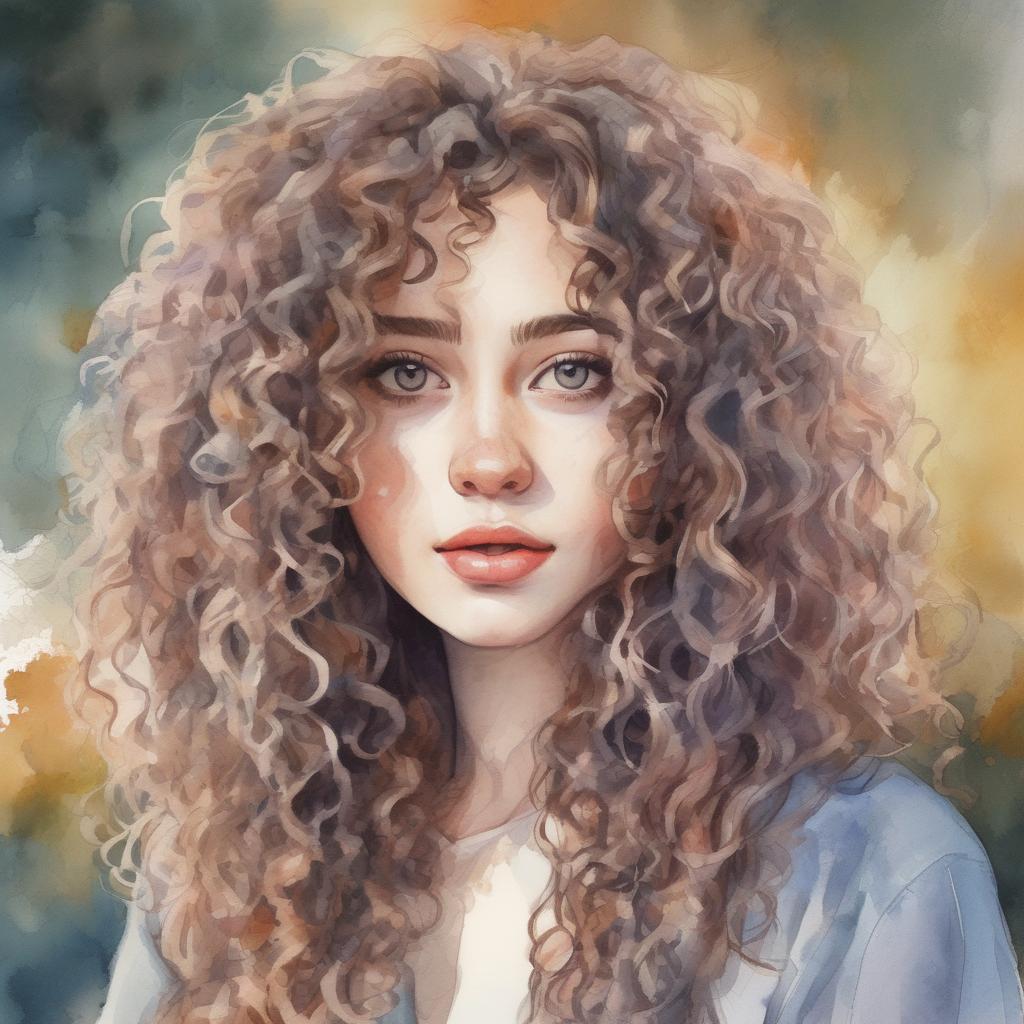}
        \caption{SDXL}
    \end{subfigure}
    \hfill
    \begin{subfigure}[b]{0.23\textwidth}
        \captionsetup{labelformat=empty}
        \centering
        \includegraphics[width=\textwidth]{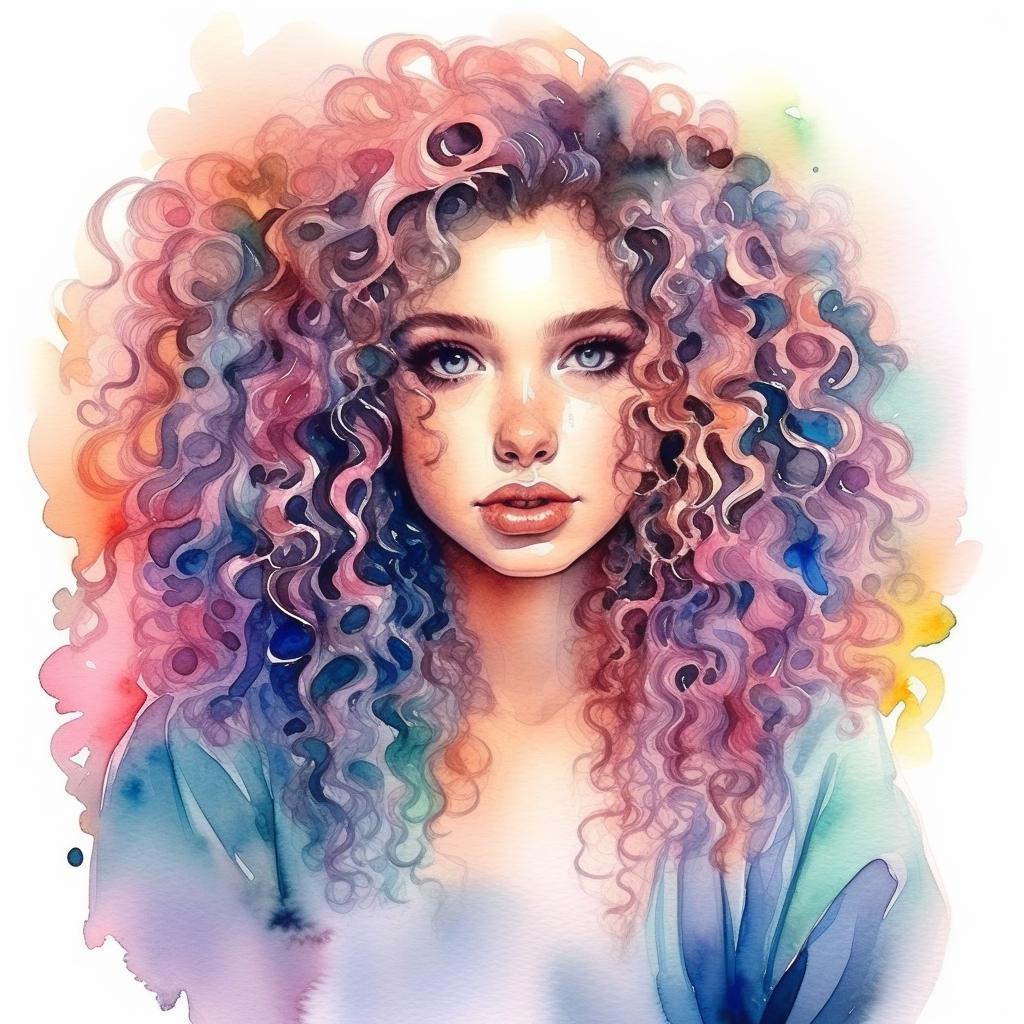}
        \caption{PixArt-Alpha}
    \end{subfigure}
    \captionsetup{labelformat=empty}
    \caption{Beautiful girl with curly hair, watercolor style, colorful, pastel colors, ultra detailed.}
    \end{subfigure}

    \vspace{0.5cm} 

    \begin{subfigure}[b]{\textwidth}
        \captionsetup{labelformat=empty}
    \begin{subfigure}[b]{0.23\textwidth}
        \captionsetup{labelformat=empty}
        \centering
        \includegraphics[width=\textwidth]{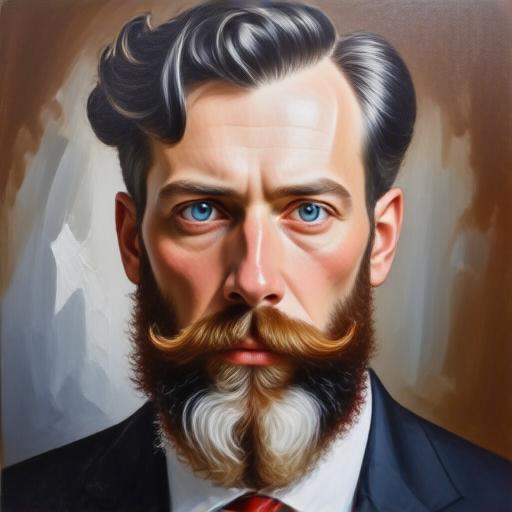}
        \caption{Ours}
    \end{subfigure}
    \hfill
    \begin{subfigure}[b]{0.23\textwidth}
        \captionsetup{labelformat=empty}
        \centering
        \includegraphics[width=\textwidth]{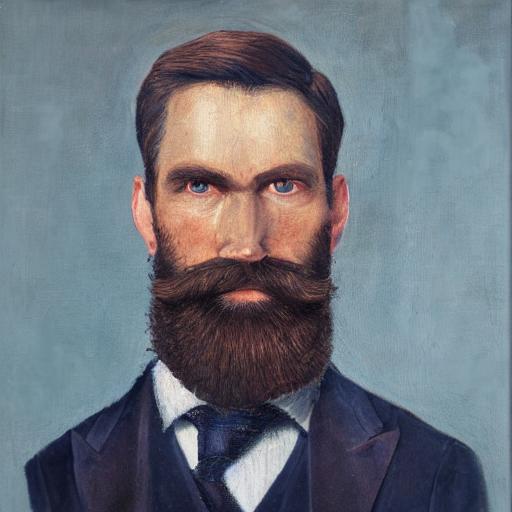}
        \caption{SDv1.5}
    \end{subfigure}
    \hfill
    \begin{subfigure}[b]{0.23\textwidth}
        \captionsetup{labelformat=empty}
        \centering
        \includegraphics[width=\textwidth]{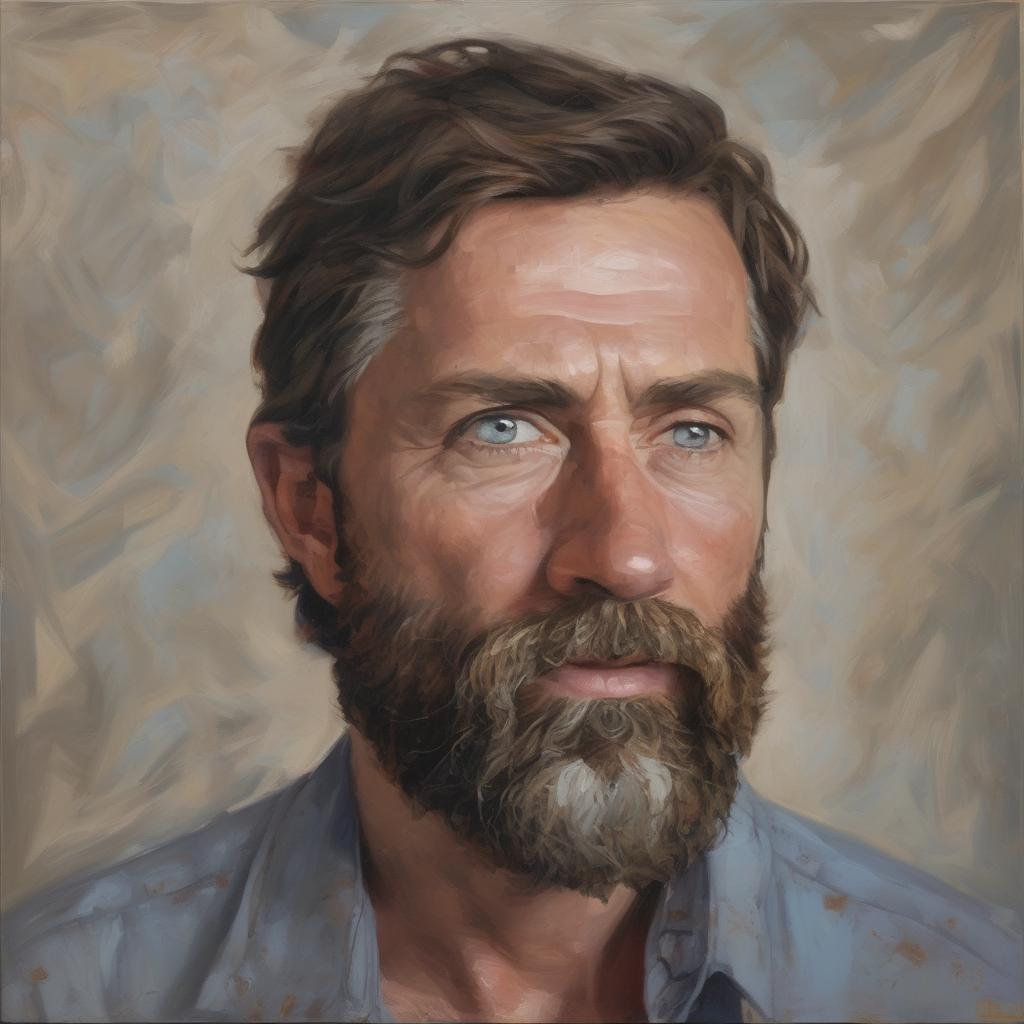}
        \caption{SDXL}
    \end{subfigure}
    \hfill
    \begin{subfigure}[b]{0.23\textwidth}
        \captionsetup{labelformat=empty}
        \centering
        \includegraphics[width=\textwidth]{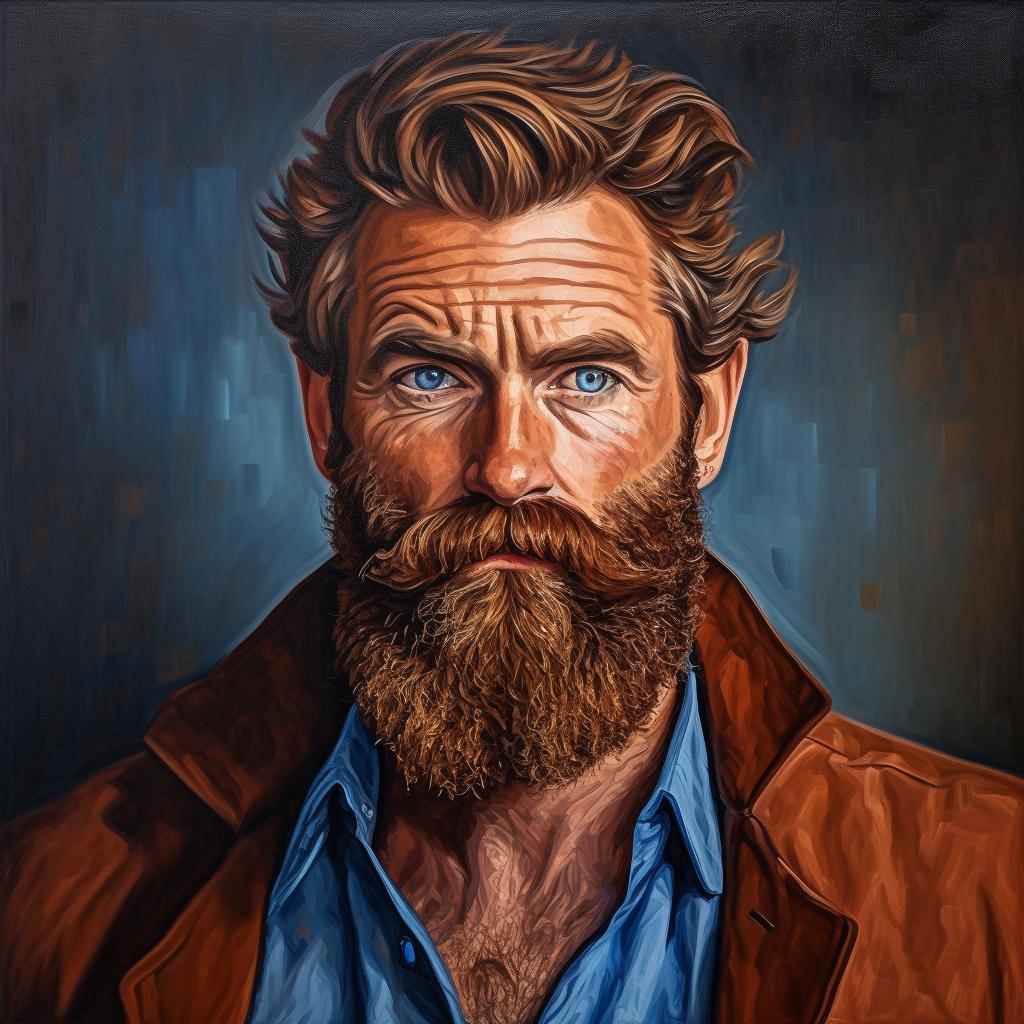}
        \caption{PixArt-Alpha}
    \end{subfigure}
    \captionsetup{labelformat=empty}
    \caption{A portrait oil painting of a smart man in his mid - 4 0 s, brown hair, blue eyes, salt and pepper beard, full beard, red skin, big eye brows, highly detailed, artistic, intricate.}
    \end{subfigure}

    \vspace{0.5cm} 

    \begin{subfigure}[b]{\textwidth}
        \captionsetup{labelformat=empty}
    \begin{subfigure}[b]{0.23\textwidth}
        \captionsetup{labelformat=empty}
        \centering
        \includegraphics[width=\textwidth]{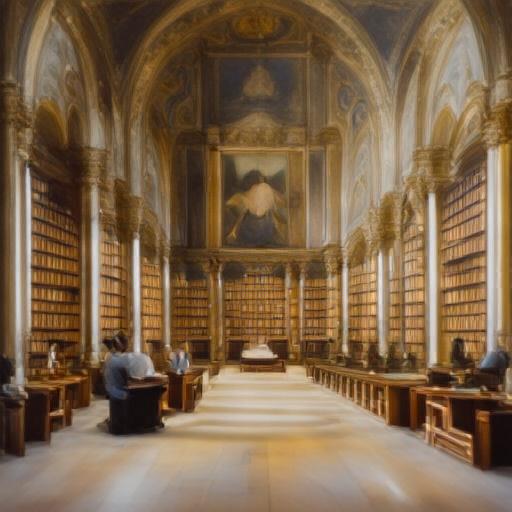}
        \caption{Ours}
    \end{subfigure}
    \hfill
    \begin{subfigure}[b]{0.23\textwidth}
        \captionsetup{labelformat=empty}
        \centering
        \includegraphics[width=\textwidth]{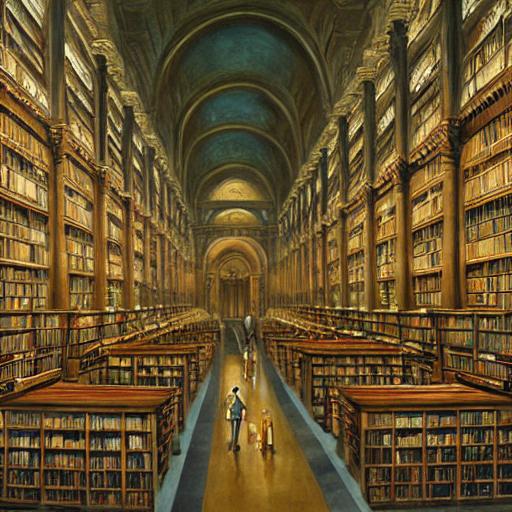}
        \caption{SDv1.5}
    \end{subfigure}
    \hfill
    \begin{subfigure}[b]{0.23\textwidth}
        \captionsetup{labelformat=empty}
        \centering
        \includegraphics[width=\textwidth]{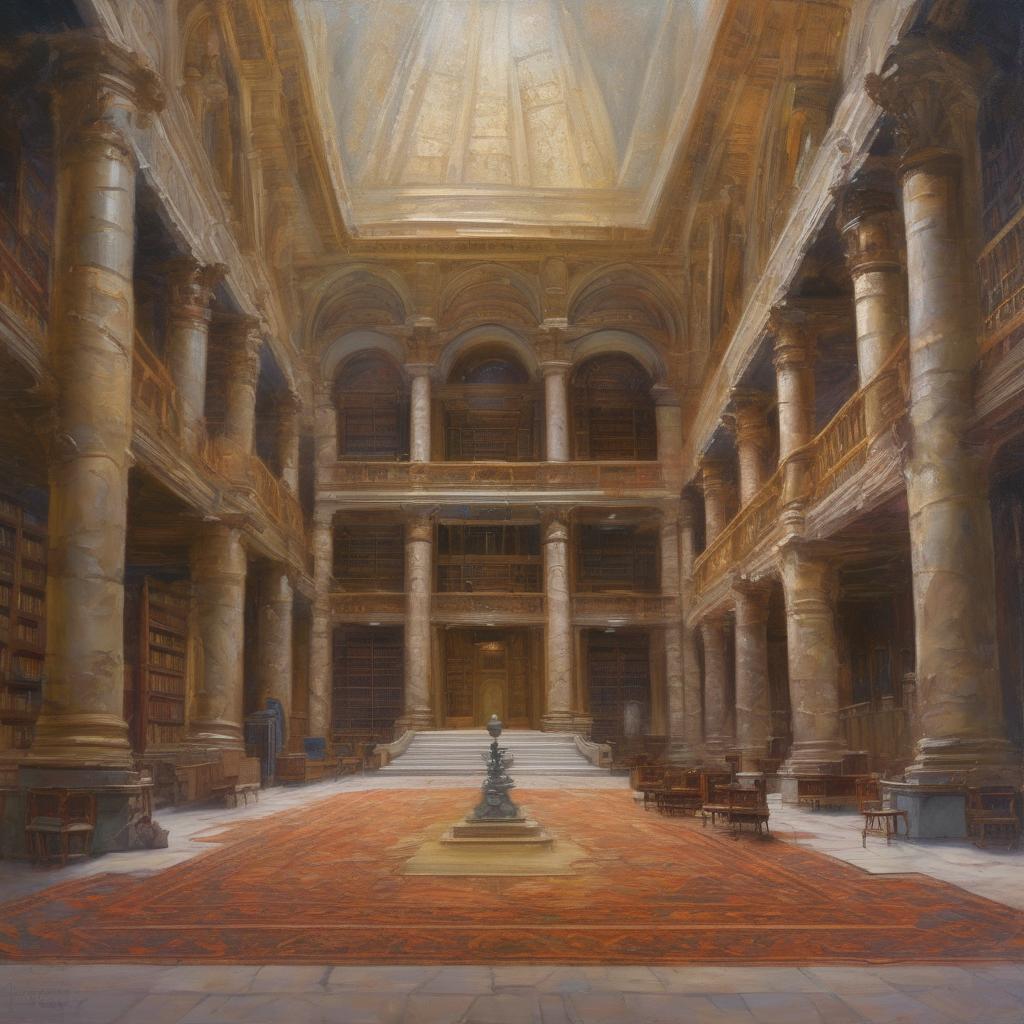}
        \caption{SDXL}
    \end{subfigure}
    \hfill
    \begin{subfigure}[b]{0.23\textwidth}
        \captionsetup{labelformat=empty}
        \centering
        \includegraphics[width=\textwidth]{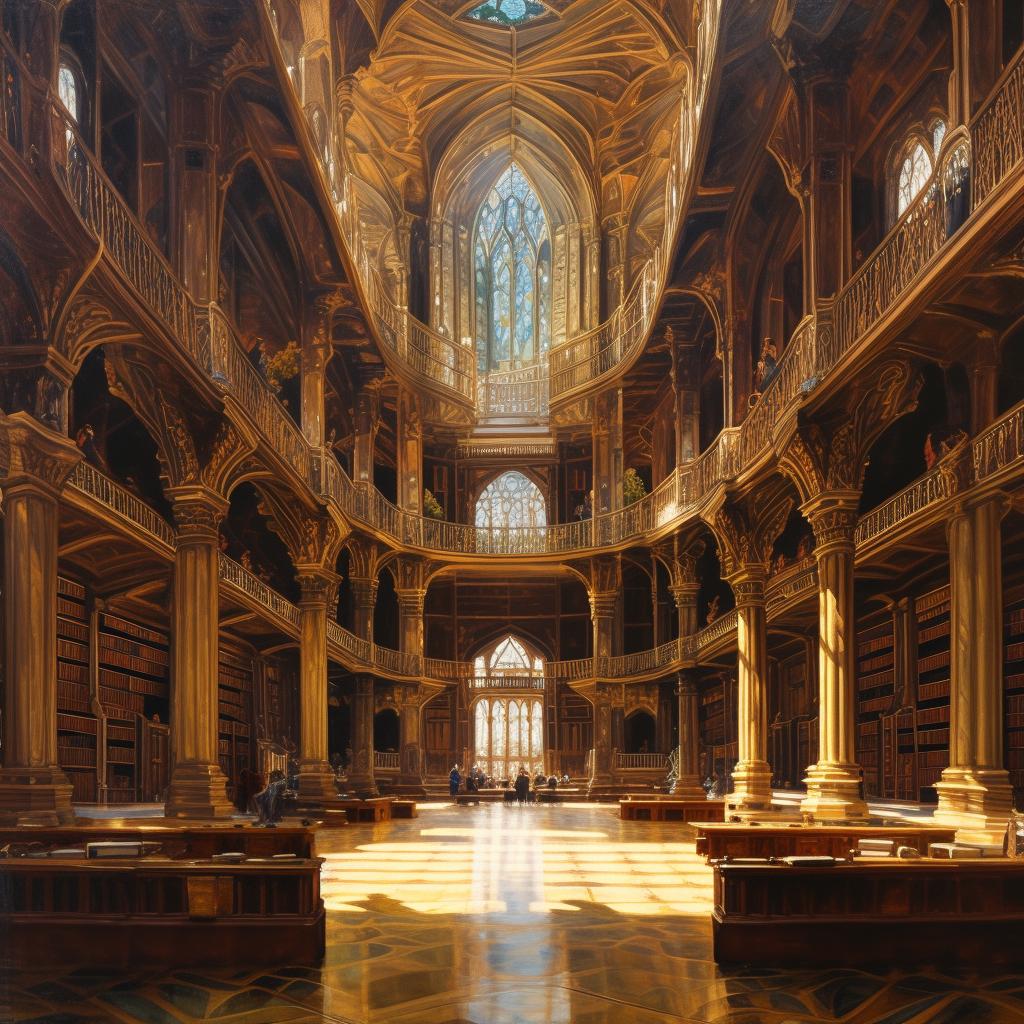}
        \caption{PixArt-Alpha}
    \end{subfigure}
    \captionsetup{labelformat=empty}
    \caption{The grand hall of the sacred library oil painting by james gurney.}
    \end{subfigure}

    \vspace{0.5cm} 

    \begin{subfigure}[b]{\textwidth}
        \captionsetup{labelformat=empty}
    \begin{subfigure}[b]{0.23\textwidth}
        \captionsetup{labelformat=empty}
        \centering
        \includegraphics[width=\textwidth]{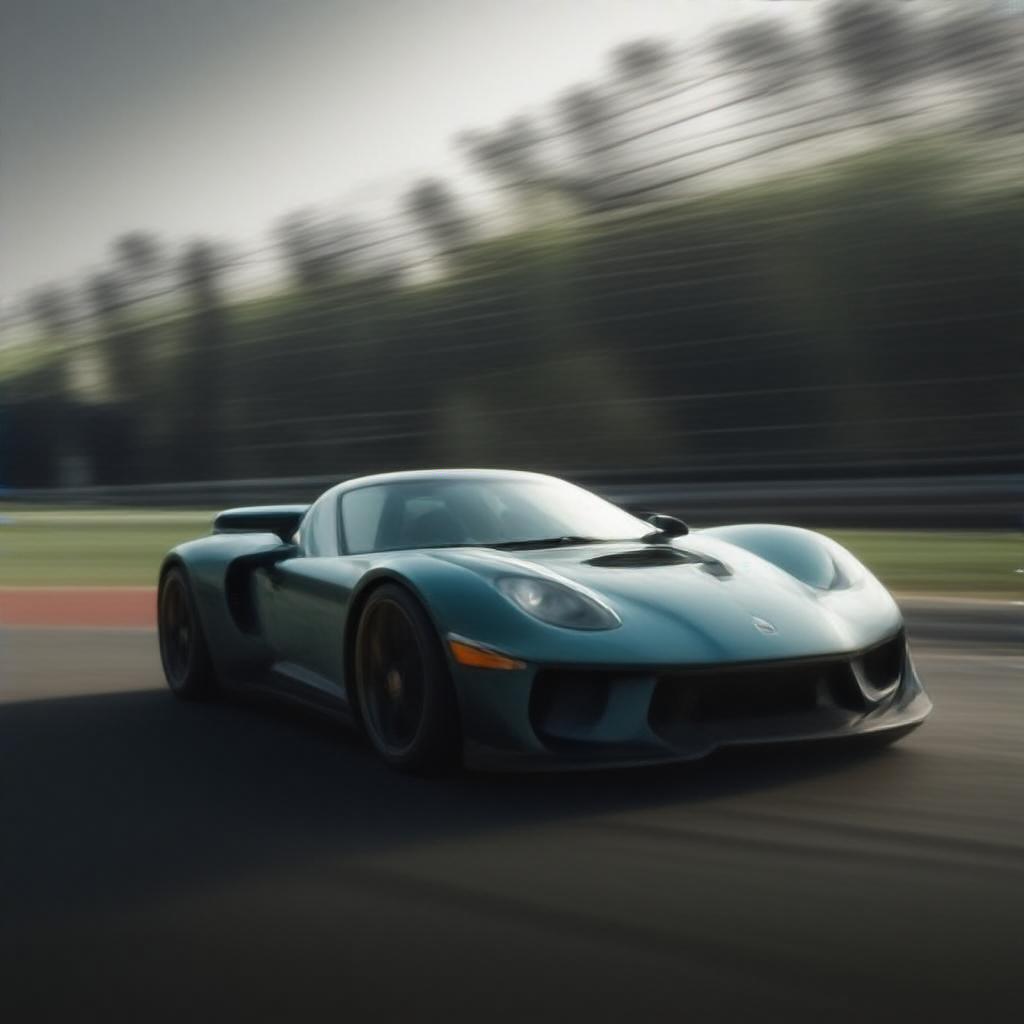}
        \caption{Ours}
    \end{subfigure}
    \hfill
    \begin{subfigure}[b]{0.23\textwidth}
        \captionsetup{labelformat=empty}
        \centering
        \includegraphics[width=\textwidth]{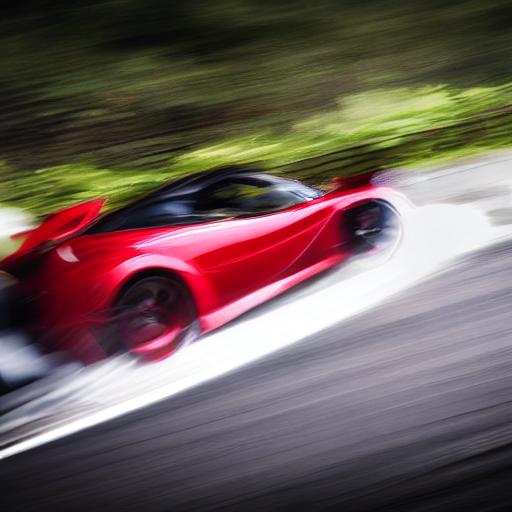}
        \caption{SDv1.5}
    \end{subfigure}
    \hfill
    \begin{subfigure}[b]{0.23\textwidth}
        \captionsetup{labelformat=empty}
        \centering
        \includegraphics[width=\textwidth]{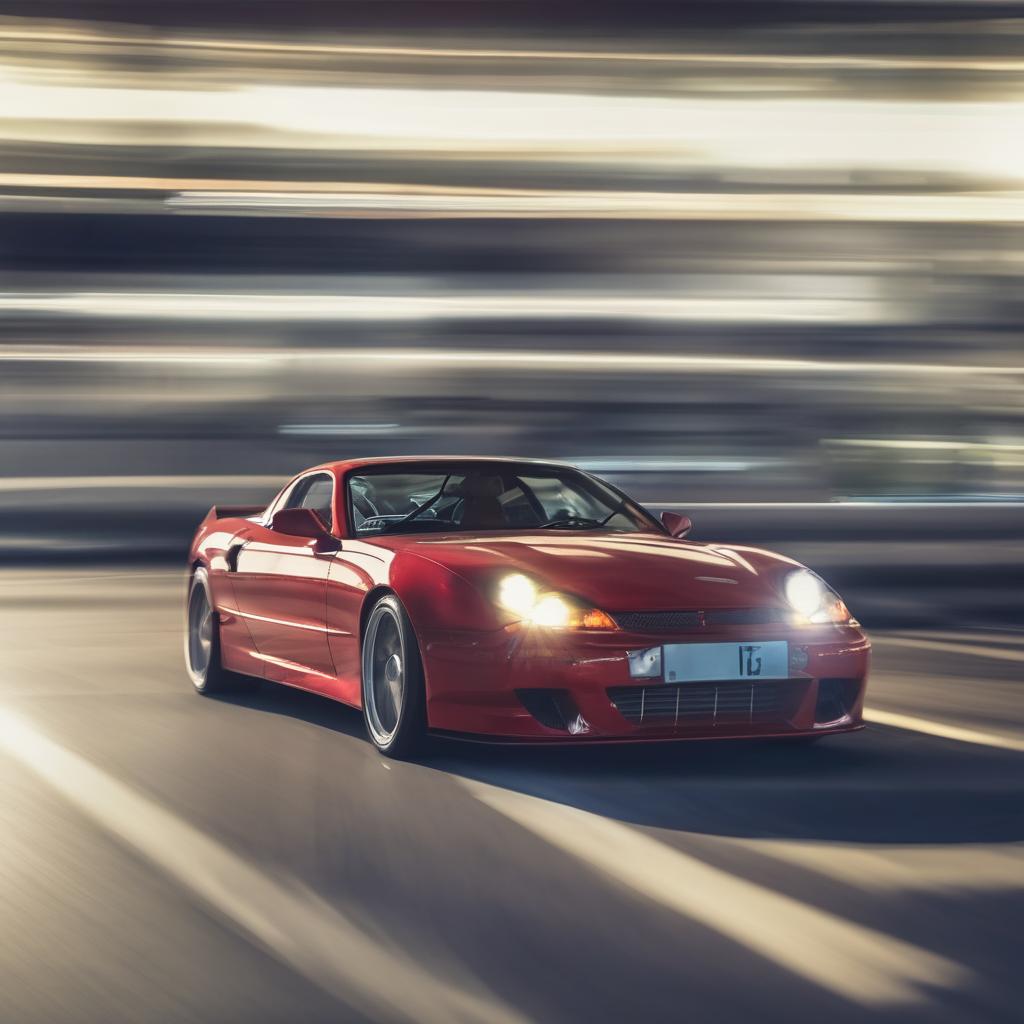}
        \caption{SDXL}
    \end{subfigure}
    \hfill
    \begin{subfigure}[b]{0.23\textwidth}
        \captionsetup{labelformat=empty}
        \centering
        \includegraphics[width=\textwidth]{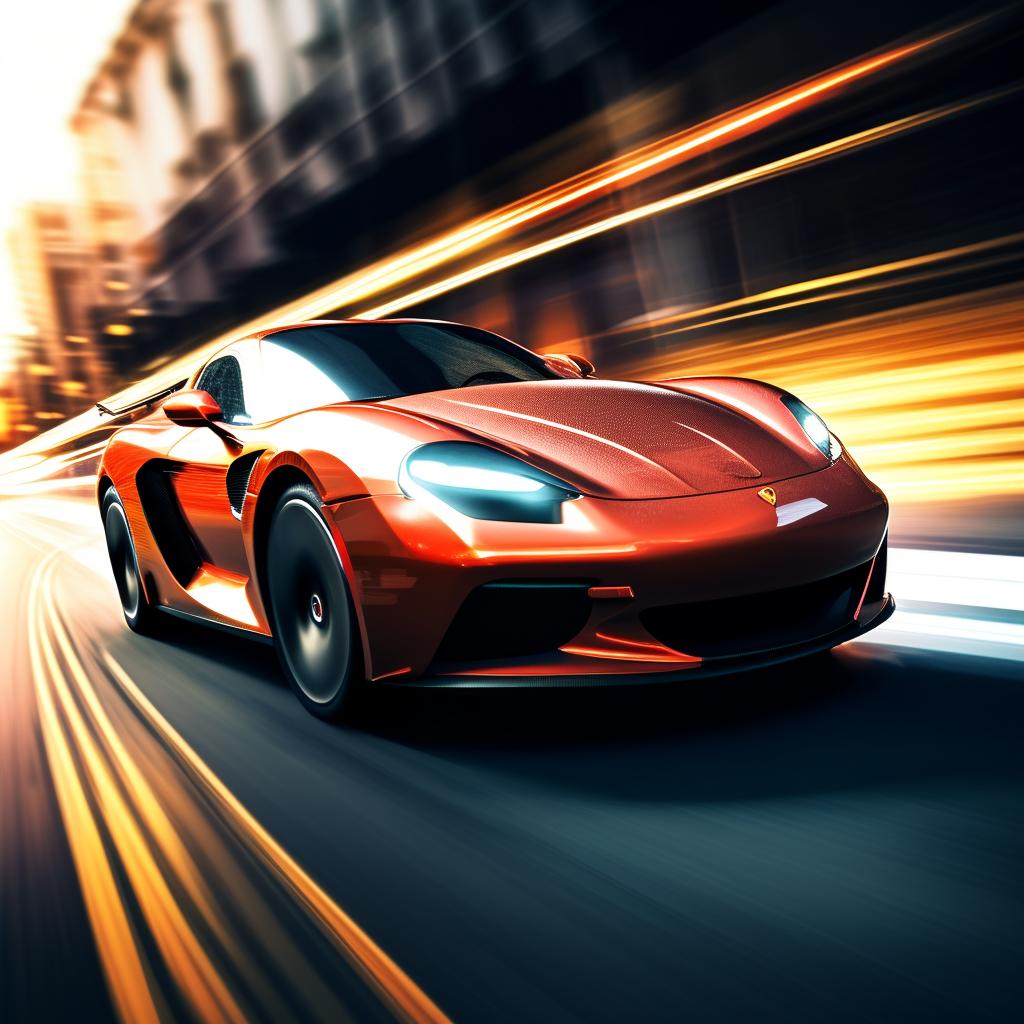}
        \caption{PixArt-Alpha}
    \end{subfigure}
    \captionsetup{labelformat=empty}
    \caption{A sports car, motion blur.}
    \end{subfigure}
    
    \caption{Side-by-side comparison of images generated by different methods.}
    \label{fig:sds2}
\end{figure}

\begin{figure}[]
    \centering

    \begin{subfigure}[b]{\textwidth}
        \captionsetup{labelformat=empty}
    \begin{subfigure}[b]{0.23\textwidth}
        \captionsetup{labelformat=empty}
        \centering
        \includegraphics[width=\textwidth]{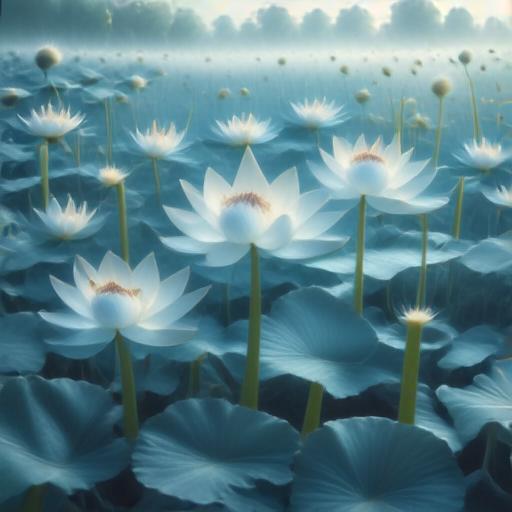}
        \caption{Ours}
    \end{subfigure}
    \hfill
    \begin{subfigure}[b]{0.23\textwidth}
        \captionsetup{labelformat=empty}
        \centering
        \includegraphics[width=\textwidth]{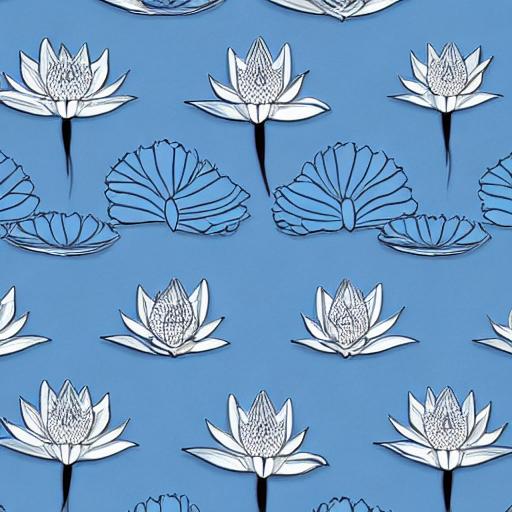}
        \caption{SDv1.5}
    \end{subfigure}
    \hfill
    \begin{subfigure}[b]{0.23\textwidth}
        \captionsetup{labelformat=empty}
        \centering
        \includegraphics[width=\textwidth]{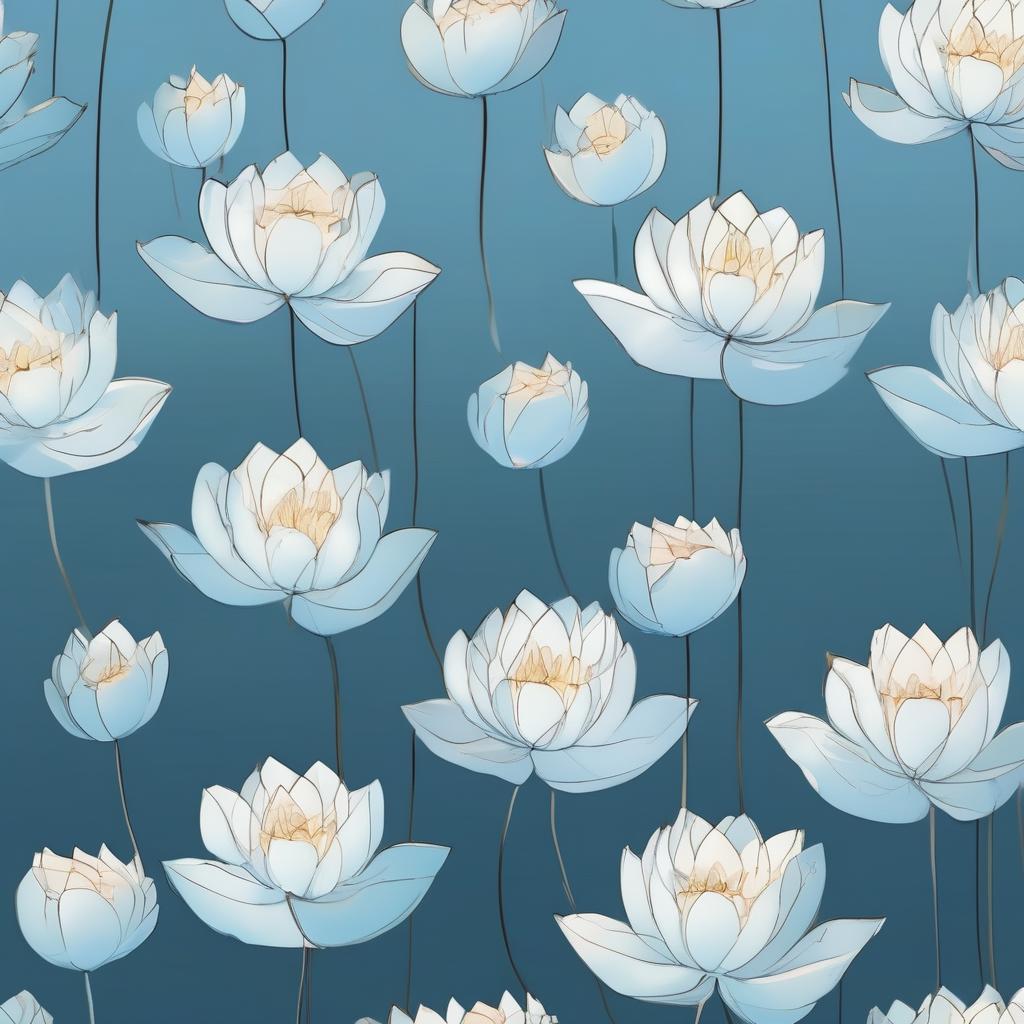}
        \caption{SDXL}
    \end{subfigure}
    \hfill
    \begin{subfigure}[b]{0.23\textwidth}
        \captionsetup{labelformat=empty}
        \centering
        \includegraphics[width=\textwidth]{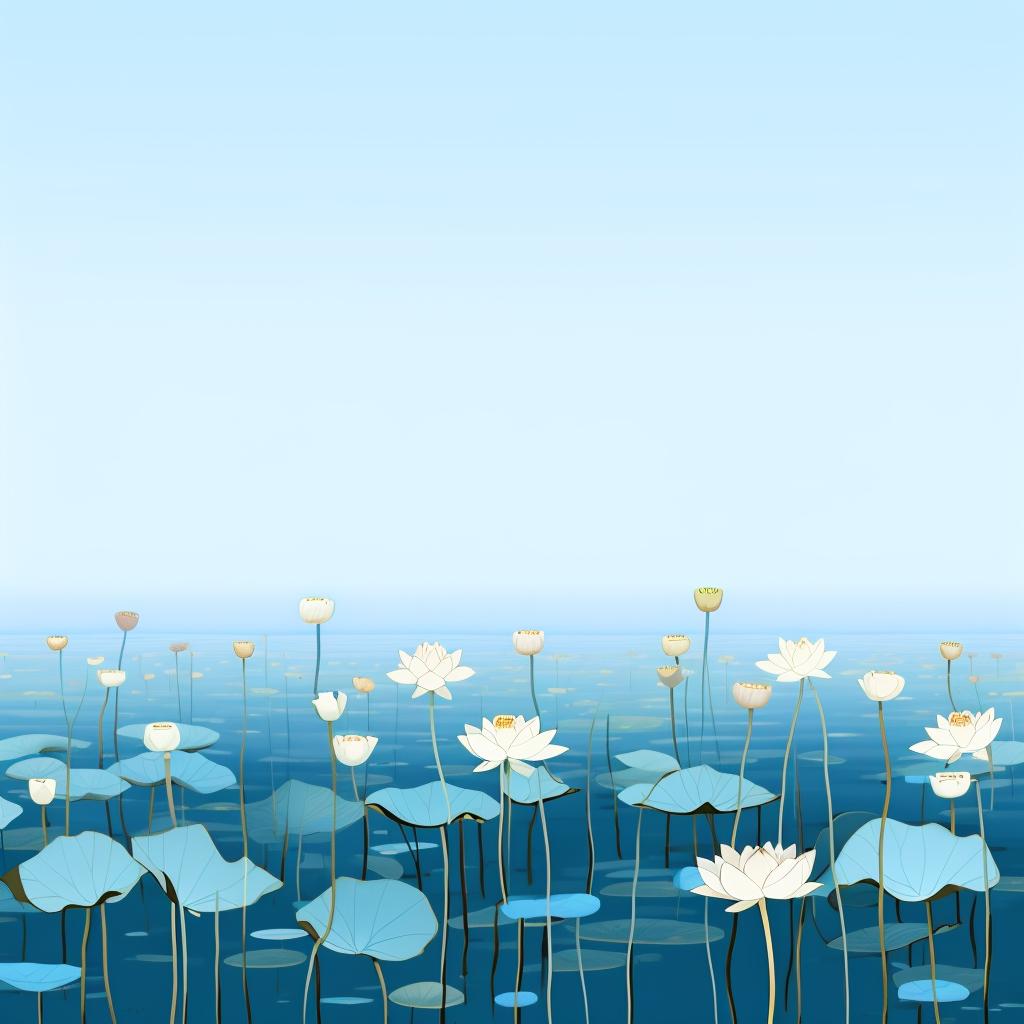}
        \caption{PixArt-Alpha}
    \end{subfigure}
    \captionsetup{labelformat=empty}
    \caption{Field of light blue lotus flowers, minimalistic art, elegant.}
    \end{subfigure}

    \vspace{0.5cm} 

    \begin{subfigure}[b]{\textwidth}
        \captionsetup{labelformat=empty}
    \begin{subfigure}[b]{0.23\textwidth}
        \captionsetup{labelformat=empty}
        \centering
        \includegraphics[width=\textwidth]{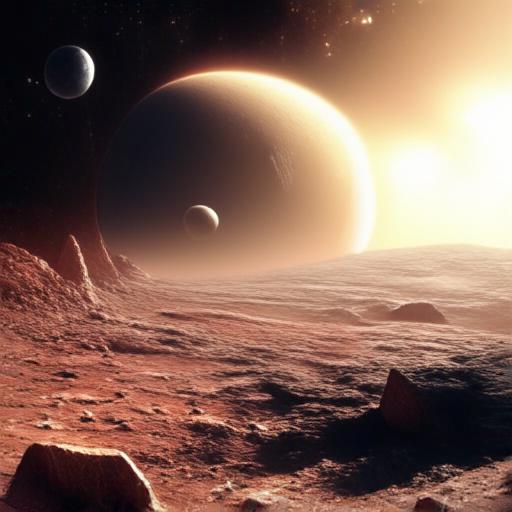}
        \caption{Ours}
    \end{subfigure}
    \hfill
    \begin{subfigure}[b]{0.23\textwidth}
        \captionsetup{labelformat=empty}
        \centering
        \includegraphics[width=\textwidth]{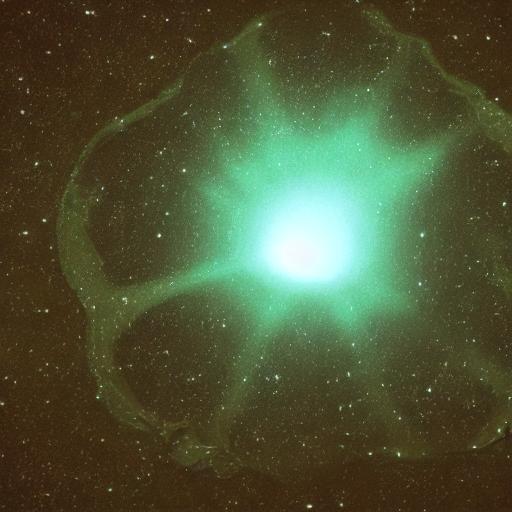}
        \caption{SDv1.5}
    \end{subfigure}
    \hfill
    \begin{subfigure}[b]{0.23\textwidth}
        \captionsetup{labelformat=empty}
        \centering
        \includegraphics[width=\textwidth]{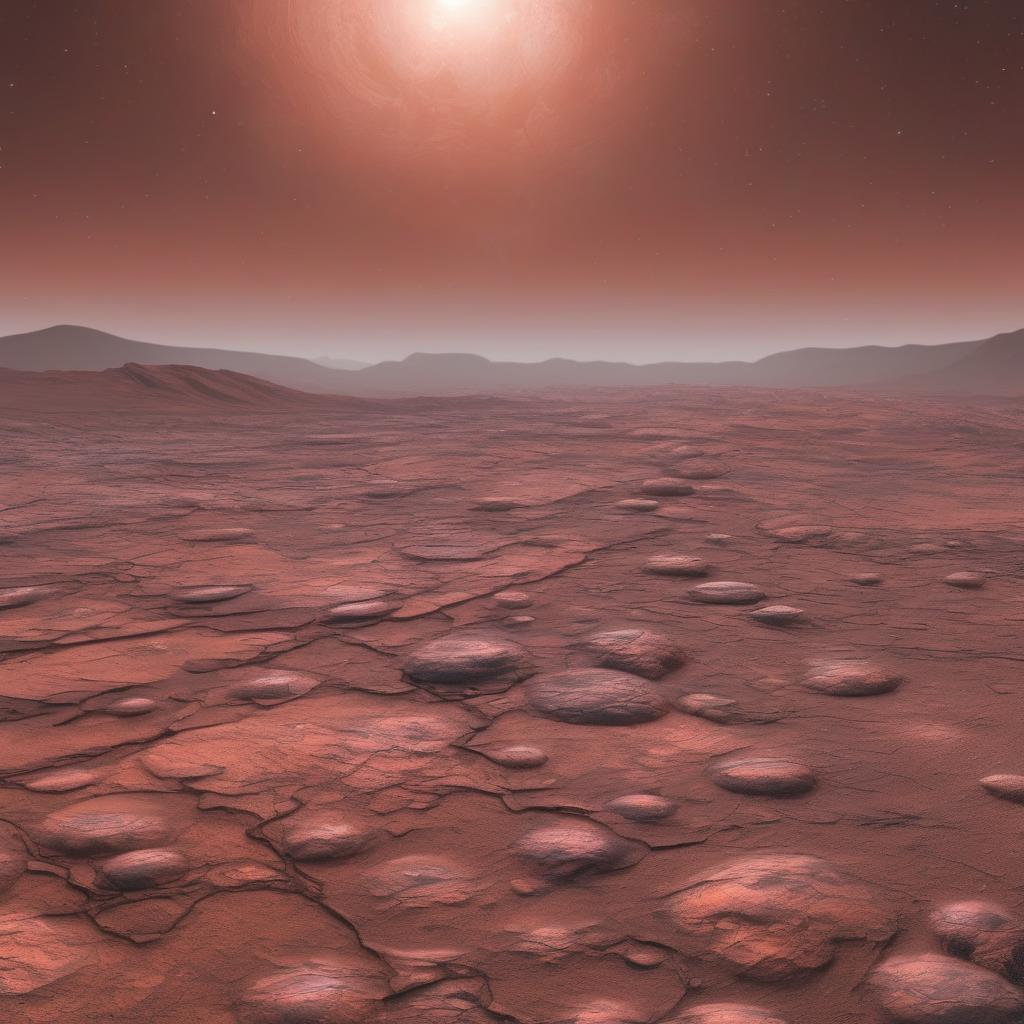}
        \caption{SDXL}
    \end{subfigure}
    \hfill
    \begin{subfigure}[b]{0.23\textwidth}
        \captionsetup{labelformat=empty}
        \centering
        \includegraphics[width=\textwidth]{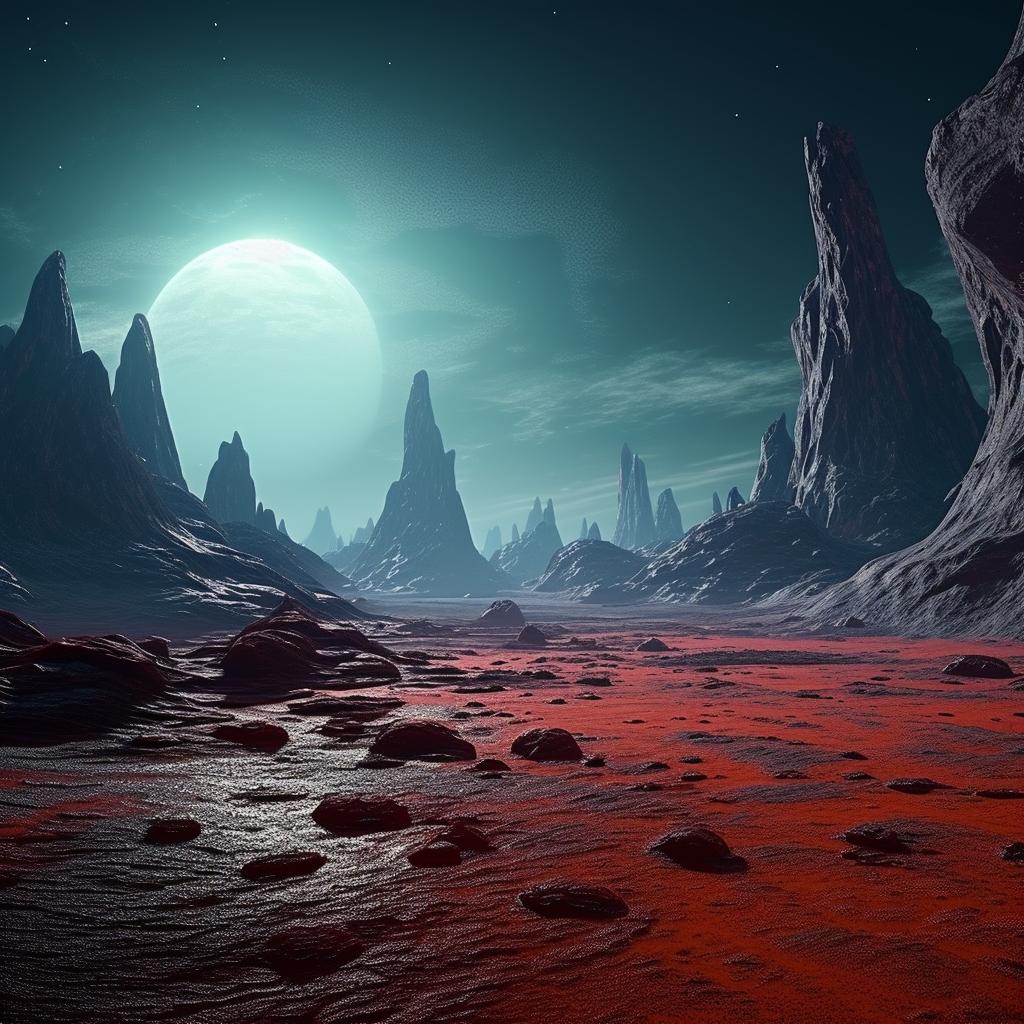}
        \caption{PixArt-Alpha}
    \end{subfigure}
    \captionsetup{labelformat=empty}
    \caption{Surface of an alien planet.}
    \end{subfigure}

    \vspace{0.5cm} 

    \begin{subfigure}[b]{\textwidth}
        \captionsetup{labelformat=empty}
    \begin{subfigure}[b]{0.23\textwidth}
        \captionsetup{labelformat=empty}
        \centering
        \includegraphics[width=\textwidth]{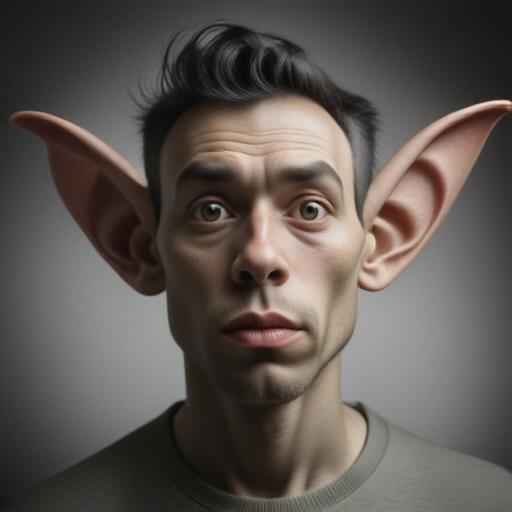}
        \caption{Ours}
    \end{subfigure}
    \hfill
    \begin{subfigure}[b]{0.23\textwidth}
        \captionsetup{labelformat=empty}
        \centering
        \includegraphics[width=\textwidth]{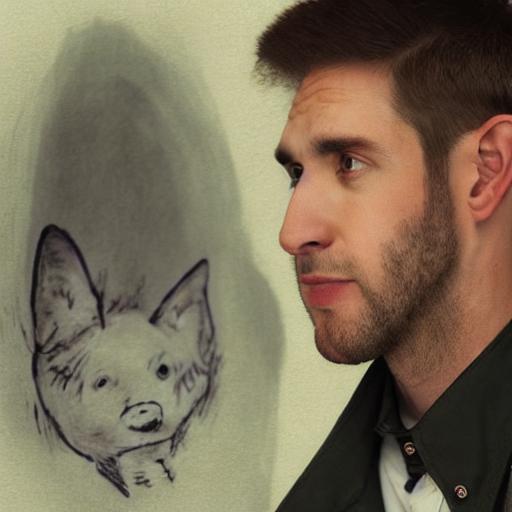}
        \caption{SDv1.5}
    \end{subfigure}
    \hfill
    \begin{subfigure}[b]{0.23\textwidth}
        \captionsetup{labelformat=empty}
        \centering
        \includegraphics[width=\textwidth]{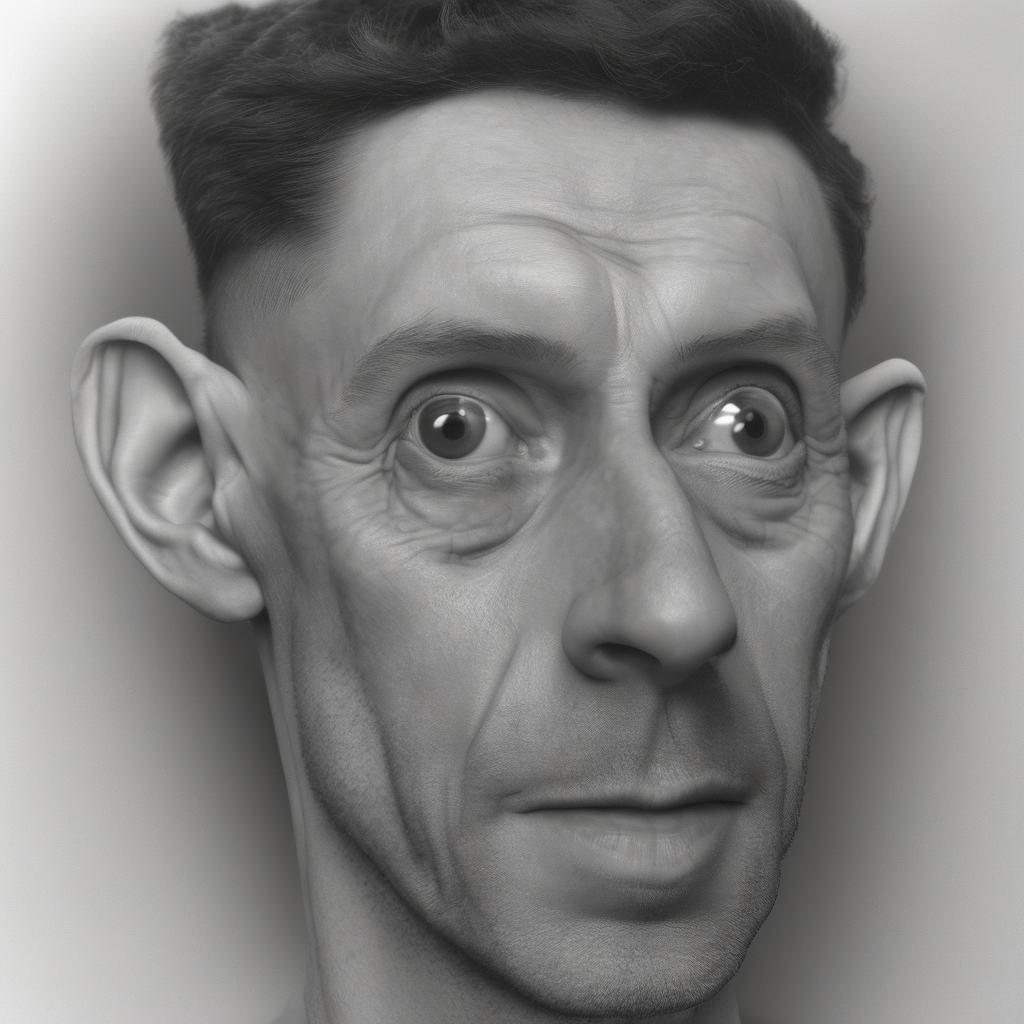}
        \caption{SDXL}
    \end{subfigure}
    \hfill
    \begin{subfigure}[b]{0.23\textwidth}
        \captionsetup{labelformat=empty}
        \centering
        \includegraphics[width=\textwidth]{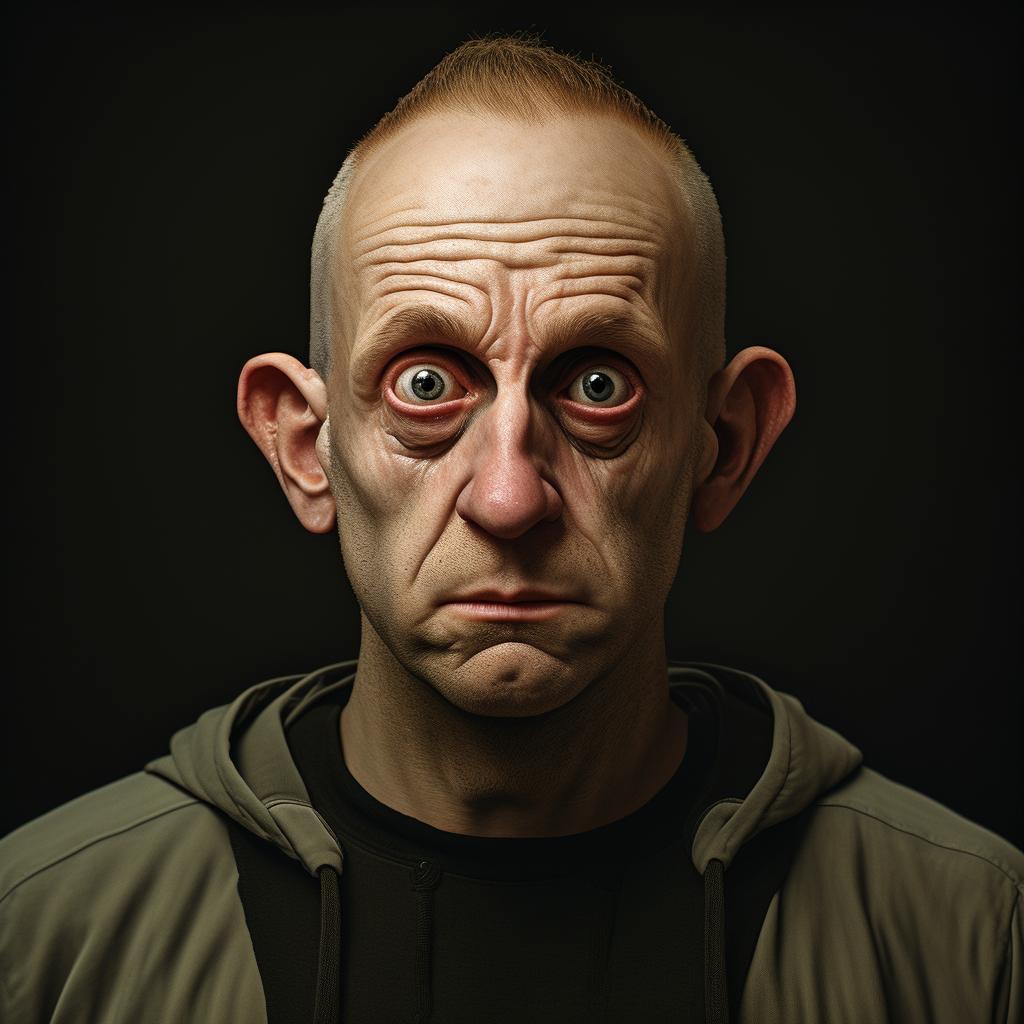}
        \caption{PixArt-Alpha}
    \end{subfigure}
    \captionsetup{labelformat=empty}
    \caption{A realistic photo of a man with big ears.}
    \end{subfigure}

    \vspace{0.5cm} 

    \begin{subfigure}[b]{\textwidth}
        \captionsetup{labelformat=empty}
    \begin{subfigure}[b]{0.23\textwidth}
        \captionsetup{labelformat=empty}
        \centering
        \includegraphics[width=\textwidth]{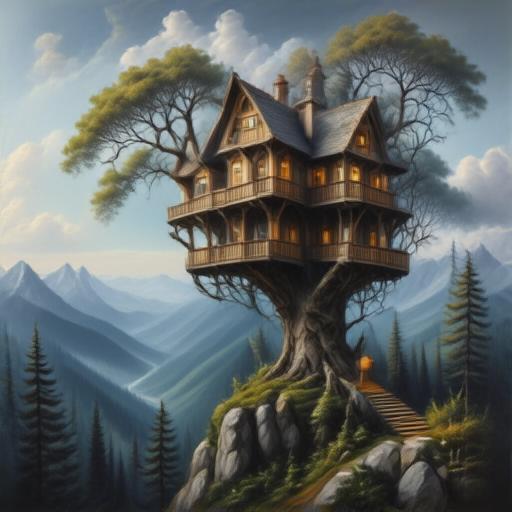}
        \caption{Ours}
    \end{subfigure}
    \hfill
    \begin{subfigure}[b]{0.23\textwidth}
        \captionsetup{labelformat=empty}
        \centering
        \includegraphics[width=\textwidth]{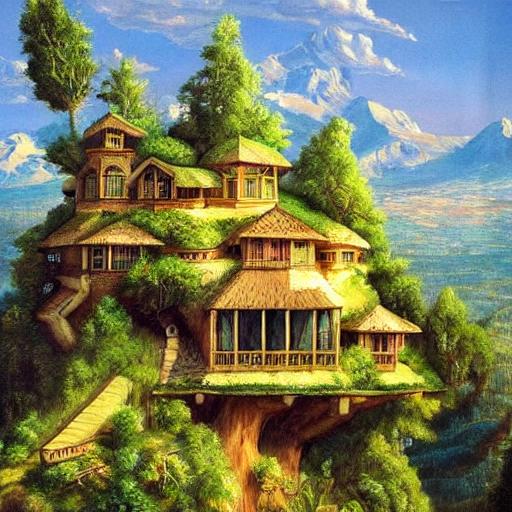}
        \caption{SDv1.5}
    \end{subfigure}
    \hfill
    \begin{subfigure}[b]{0.23\textwidth}
        \captionsetup{labelformat=empty}
        \centering
        \includegraphics[width=\textwidth]{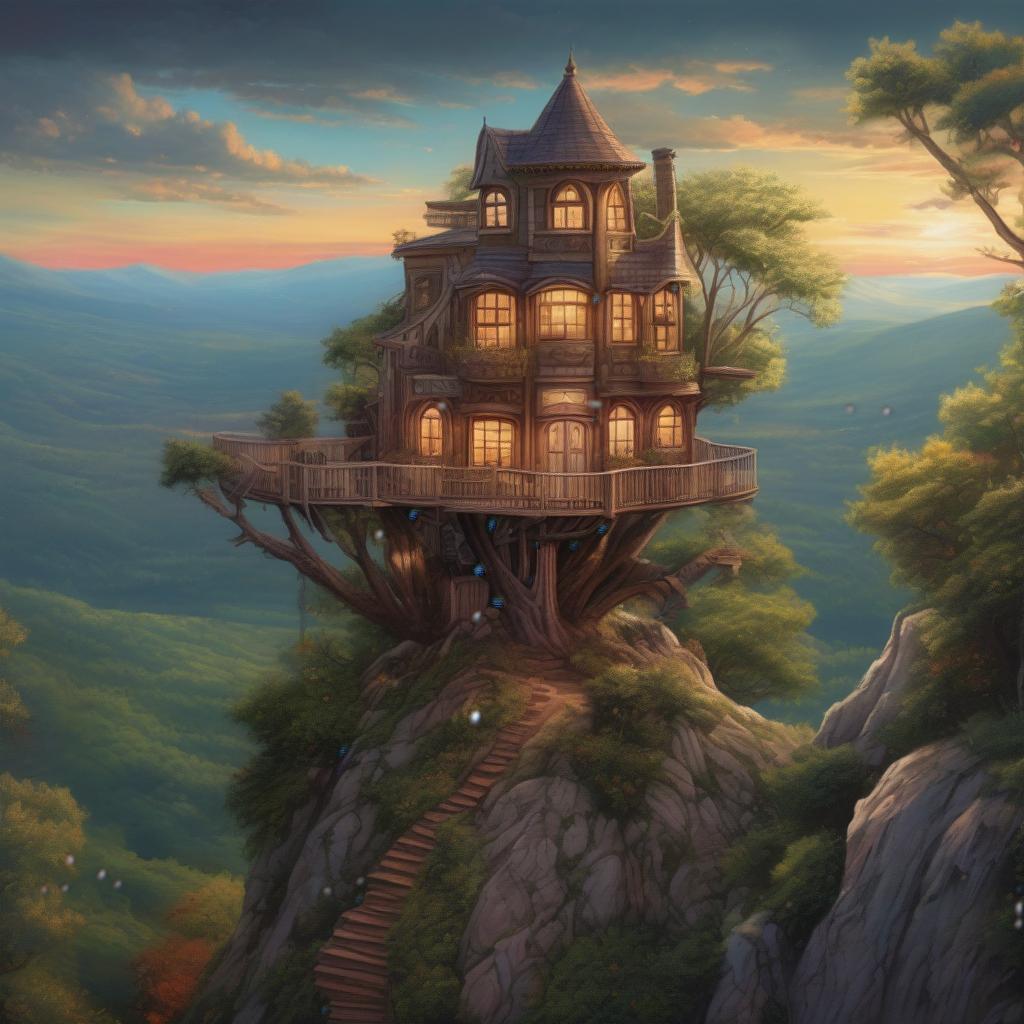}
        \caption{SDXL}
    \end{subfigure}
    \hfill
    \begin{subfigure}[b]{0.23\textwidth}
        \captionsetup{labelformat=empty}
        \centering
        \includegraphics[width=\textwidth]{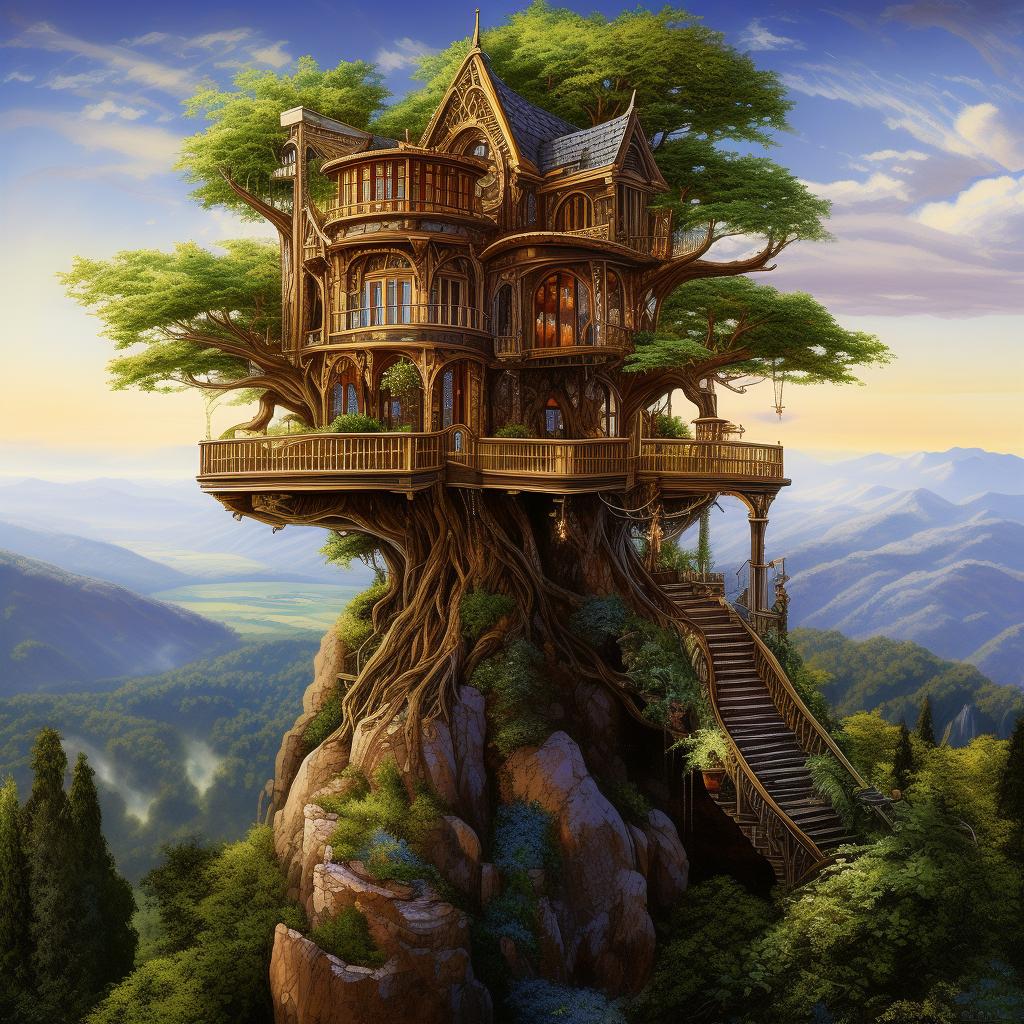}
        \caption{PixArt-Alpha}
    \end{subfigure}
    \captionsetup{labelformat=empty}
    \caption{Fancy treehouse mansion on top of a mountain overlooking a view of the valley magical realism detailed painting.}
    \end{subfigure}
    
    \caption{Side-by-side comparison of images generated by different methods.}
    \label{fig:sds3}
\end{figure}

In this section, we provide side-by-side examples of images generated by SDv1.5, SDXL, PixArt-Alpha, and our models for comprehensive comparisons in Figure~\ref{fig:sds1}, Figure~\ref{fig:sds2}, and Figure~\ref{fig:sds3}. The evaluation prompts are randomly sampled from the benchmark prompts of ImageReward \cite{xu2023imagereward}.

\end{document}